\documentclass[journal]{IEEEtran}
\usepackage{graphicx}
\usepackage{amsmath,amssymb} 
\usepackage{color}
\usepackage{multirow}
\usepackage{caption}
\usepackage{xparse}
\usepackage{tabularx}
\newcommand*{\affmark}[1][*]{\textsuperscript{#1}}
\makeatletter
\newcommand*{\rom}[1]{\expandafter\@slowromancap\romannumeral #1@}
\makeatother
\ifCLASSINFOpdf
\else
\fi

\begin{document}
%
\title{Drivers Drowsiness Detection using Condition-Adaptive Representation Learning Framework}
%
%
%

\author{Jongmin Yu $^{1}$, Sangwoo Park $^{1}$,Sangwook Lee $^{2}$,~\IEEEmembership{Members,~IEEE,}\\
        and Moongu Jeon $^{1,}$*,~\IEEEmembership{Senior Member,~IEEE,}
\thanks{$^{1}$ \quad School of Electrical Engineering and Computer Science, Gwangju Institute of Science and Technology (GIST), Gwangju, 61005, South Korea; jm.andrew.yu@gmail.com, \{parkswoo,mgjeon\}@gist.ac.kr}
\thanks{$^{2}$ \quad Department of Information Communication Engineering, Mokwon University, Daejeon, 35349, South Korea; \affmark[3]slee@mokwon.ac.kr}
\thanks{$^{*}$ \quad represents the corresponding author}
\thanks{Manuscript received April 19, 2005; revised August 26, 2015.}}

\markboth{Journal of \LaTeX\ Class Files,~Vol.~14, No.~8, August~2015}%
{Shell \MakeLowercase{\textit{et al.}}: Bare Demo of IEEEtran.cls for IEEE Journals}

\maketitle

\begin{abstract}	
We propose a condition-adaptive representation learning framework for the driver drowsiness detection based on 3D-deep convolutional neural network. The proposed framework consists of four models: spatio-temporal representation learning, scene condition understanding, feature fusion, and drowsiness detection. The spatio-temporal representation learning extracts features that can describe motions and appearances in video simultaneously. The scene condition understanding classifies the scene conditions related to various conditions about the drivers and driving situations such as statuses of wearing glasses, illumination condition of driving, and motion of facial elements such as head, eye, and mouth. The feature fusion generates a condition-adaptive representation using two features extracted from above models. The detection model recognizes drivers drowsiness status using the condition-adaptive representation. The condition-adaptive representation learning framework can extract more discriminative features focusing on each scene condition than the general representation so that the drowsiness detection method can provide more accurate results for the various driving situations.  The proposed framework is evaluated with the NTHU Drowsy Driver Detection video dataset. The experimental results show that our framework outperforms the existing drowsiness detection methods based on visual analysis.
\end{abstract}


\IEEEpeerreviewmaketitle

\section{Introduction}

\IEEEPARstart{D}{r}iver drowsiness detection is one of the essential functions in the advanced driver assistant systems (ADAS) for preventing fatal accidents from the people on a road. Many drivers and pedestrians are killed or significantly injured by drowsy driving. The report of the National Sleep Foundation’s Sleep in America poll presents 60\% of Americans have an experience of drowsiness driving, and 37\% have experienced falling asleep while driving in the recent one year. According to the report of the national highway traffic safety administration in the USA, the driver fatigue is closely related to the 100,000 of car crashes reported by polices. By this report, this car crashes made 1,550 deaths, 71,000 injuries, and 12.5 billion in monetary losses \cite{schroeder2013national}. Also, the car crash by the driver drowsiness is not unique to drivers in the USA, drowsiness contributes to as many as 7\% of crashes in the United Kingdom and 3.9\% of crashes in Norway\cite{maycock1996sleepiness, sagberg1999road}. The majority of drowsiness-related car accidents, approximately 80\%, might be classified as individual vehicle run off road crashes, where a driver lost the controlling their vehicle and eventually departed their lane or smashed into the rear of the car ahead \cite{pack1995characteristics}. These figures may be the tip of the iceberg because of not only it is hard to attribute the cause of crashes to drowsiness but also the criteria for recognizing drowsiness differ depending on the driver \cite{schroeder2013national}. There is no Breathalyzer equivalent for drowsiness. Therefore, in order to prevent these losses of life and property, it is an important challenge to develop a driver drowsiness detection method.

The approaches for driver drowsiness detection could be classified based on their target domain to analysis. One approach is to directly analyze the driver's behaviour to identify changes in driver behaviour. This approach analyzes facial elements such as eye and mouth using visual sensors \cite{garcia2012vision, mbouna2013visual, wang2012method, minkov2012comparison, panning2011color, kurylyak2012detection, suzuki2006measurement}, or detects particuar patterns in electrophysiological signals occurring when a driver is falling asleep \cite{khushaba2011driver, patel2011applying, tran2010improving, papadelis2007monitoring}. Other approaches indirectly infer a driver's state through analysis of signals extracted from the steering system \cite{ersal2010model, yang2009detection, liu2009predicting, takei2005estimate, wakita2006driver}.

The most commonly applied and theoretically rigorous approach involves the analysis of electrical bio-signals e.g., electroencephalogram (EEG) or facial elements such as eye based on percent eye-closure over a fixed time window (PERCLOS) \cite{dinges1998perclos}. Dinges et al. had verified that the approach using PERCLOS had over than 90\% accuracy in recognizing degraded performance during a vigilance task. This figure demonstrated that the PERCLOS was more reliable across drivers than EEG, blinks, and head position in the study \cite{dinges1998perclos}. Khushaba et al. proposed the driver drowsiness detection method which employs fuzzy mutual-information-based wavelet packet transform model for extracting drowsiness-related information from a set of EEG, electrooculogram (EOG), and electrocardiogram (ECG) signals \cite{khushaba2011driver}. Papadelis et al. developed drowsiness monitoring system using onboard electrophysiological recording systems \cite{papadelis2007monitoring}. Aforementioned methods identify the change of patterns of signals such as brain activity or heartbeat to measure the strength of fatigue of drivers. These signals reflect brain electrical activity and can provide more discriminative information than other features in analyzing the driver's conditions. For these reasons, the methods using biomedical signals captured from drivers had provided relatively higher accurate detection results than other methods based on visual analysis or measuring the steering signals. Nevertheless, the main disadvantage of these methods is that the sensing equipment for the physiological signals such as EEG, ECG, and EOG, must be attached to the driver's body. The attachment of those sensors could cause inconvenience to drivers when they are driving. Additionally, the high price of sensors is one reason that they can not be used in a practical drowsiness detection system.

In addition to the methods of directly recognizing the drivers' condition through the analysis of biomedical signals, the approaches based on visual analysis of facial elements generally employ computer vision techniques such as object detection and tracking to find the interesting objects such as eye or mouth, on the image containing the driver's face  \cite{garcia2012vision, mbouna2013visual, wang2012method, minkov2012comparison, panning2011color, kurylyak2012detection, suzuki2006measurement}. Garcia et al. proposed a system which consist of three steps \cite{garcia2012vision}. Their system initially detects and tracks face and eye, and then to stabilize the performance of analyzing the status of the eye in various illumination conditions, the system conducts image filtering. This system evaluates the closure status of the eye using PERCLOS measurement. Mbouna et al. provided the analysis method for a visual feature to understand the closure state  and head pose. The proposed method monitors a driver using a single camera without any source of light \cite{mbouna2013visual}. Wang et al. presented a solution for the situation that driver is wearing glasses by combining two analysis methods for the status of eye and mouth \cite{wang2012method}. The method proposed by Dwivedi et al. extracts features using a convolutional neural network and detects eye blinking, eye closure, and yawning \cite{dwivedi2014drowsy}. Generally, these methods assume that facial expressions of extremely tired drivers, such as eye blinking, yawning, and eye and head moving, are different from facial expressions represented when drivers are not tired. These approaches classify the driver's condition as whether he/she is asleep or not, using the hand-crafted features such as the histogram of gradient (HoG) \cite{dalal2005histograms} and Haar-like features \cite{lienhart2002extended}. To extract these facial feature information, visual sensors like an RGB camera or an active infrared sensor should be installed on the vehicle dashboard, sun visor, or overhead console for taking face images of drivers. However, despite the convenience of installation, the methods based on video analysis using visual sensors solely, provide unstable detect results in many situations. For example,  general cameras cannot capture clear images at night without illumination system. The development of the drowsiness detection method using visual analysis, invariant to the light condition is still an open question.

The limitations of the above-mentioned approaches have led researchers to attend to the signals from a steering system such as the deflection of the top of the wheel from the zero point \cite{mcdonald2018contextual}. These signals are similar to electrical bio-signals in that they require significant pre-processing and transformation before they become viable input measures \cite{sayed2001unobtrusive}. Sayed and Eskandarian proposed a steering-wheel angle based method that filtered raw information for steering angle for the elimination of road curvature events, and then discretized into binary signals to represent steering patterns \cite{sayed2001unobtrusive}. This method detected the drowsiness of drivers with nearly 90\% accuracy. Similarly, Krajewski et al. presented an approach to process raw steering-wheel angle data into features represented by the signal in the time and frequency domains \cite{krajewski2009detecting}. Ersal et al. presented an approach to recognition of driving behaviours \cite{ersal2010model}, which is based on support vector machines (SVM) \cite{hearst1998support}. This approach systemically assists determination of whether a driver is asleep or not by interpreting behaviours of drivers using the linear discriminative model. Takei et al.  \cite{takei2005estimate} estimated a driver's fatigue by analyzing steering motions with the fast Fourier transform (FFT) and Chaos characteristics. These methods judge whether a driver is falling into a drowsy state by analyzing signals such as variation of velocity, acceleration, breaking, and gear change, that are recorded from the sensors embedded in steering systems. These methods are not focused on the detection of driver drowsiness directly. They try to recognize the unstable vehicle movements that are caused by various intrinsic and extrinsic reasons from analyzing steering signals. Consequently,  it can provide a more flexible system to detect unstable movements than other systems which are only focused on the detection of driver drowsiness. However, many automobile manufacturers in the world embed a particular steering system in their vehicles. In addition, these signals cannot be a clear basis to distinguish whether a driver is sleepy or not since every driver has not only a different personality but also a different driving habit.


\begin{figure}
	
	\centering
	
	\includegraphics[width=\columnwidth]{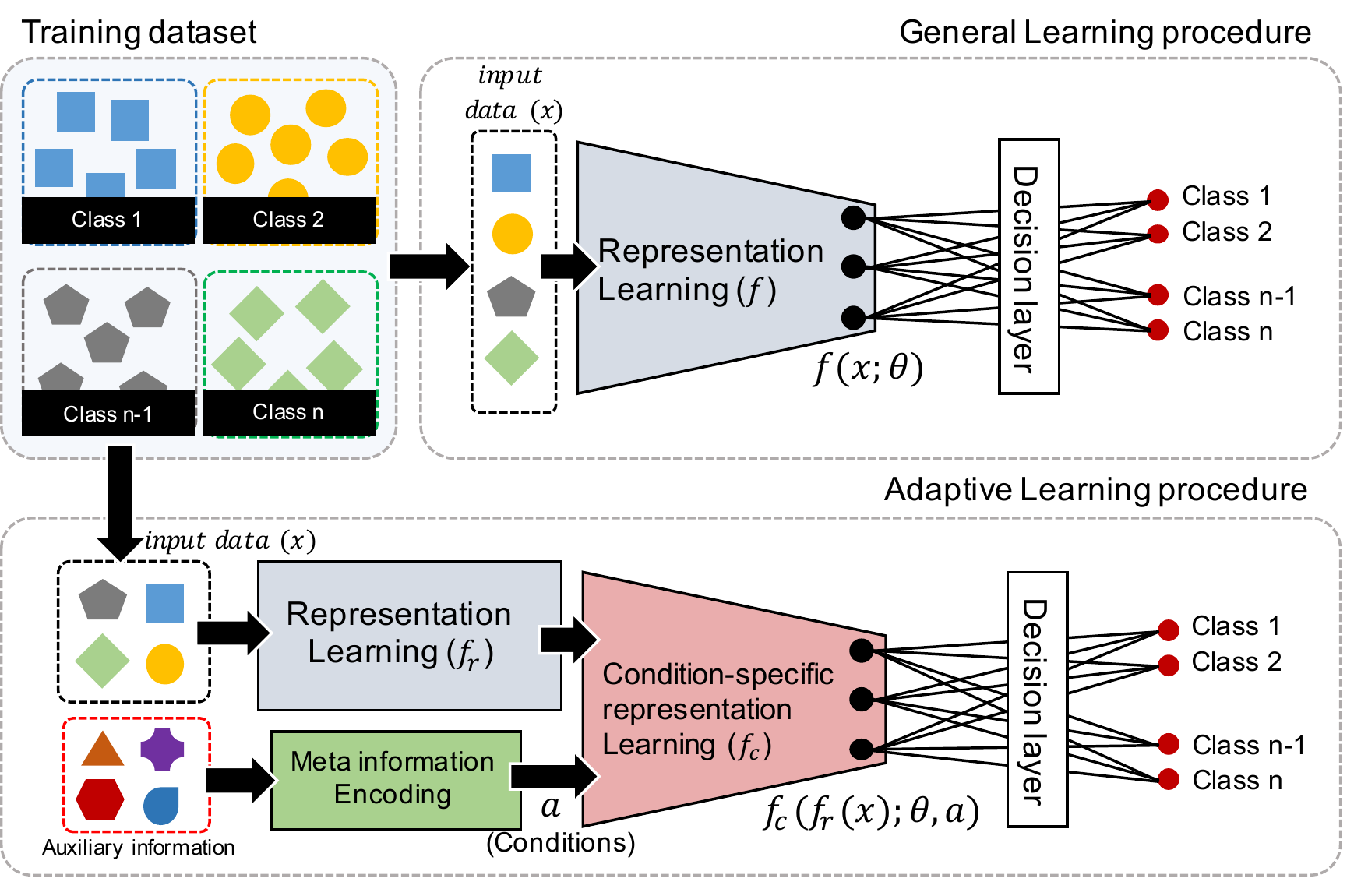}
	
	\caption{Illustrations of the processes of general represenration learning and adaptive representation learning on a classification task}
	
	\label{fig:example}
	
\end{figure}

Recently, deep learning architectures have been successfully used to solve various computer vision problems, such as image recognition \cite{simonyan2014very,girshick2016region}, object detection \cite{erhan2014scalable,ren2015faster}, gesture recognition \cite{molchanov2016online}, image segmentation \cite{qi2016dynamic}, and action recognition \cite{simonyan2014two, du2015hierarchical}. In particular, the deep learning methods \cite{simonyan2014two, du2015hierarchical} show good performance in analyzing video streams to recognize specific actions when compared with conventional methods based on hand-crafted features \cite{wang2013action,jiang2015human}. Although various methods \cite{wang2013action,jiang2015human,jain2013better} to extract superior hand-crafted features have been proposed, the key to these successes is a rich and discriminative representation extracted from multi-layer nonlinear systems in the deep learning approaches \cite{xu2015learning}. We had adopted the convolutional neural network (CNN) and multi-layer fully connected neural network (a.k.a., deep neural network) to discover significant time-space features, and showed the possibility of the deep learning method for drowsiness detection in previous works \cite{yu2016representation}. In our previous works, we had proposed the driver drowsiness detection method exploiting extra scene condition prediction to improve discriminative properties of learnt representation. However, despite outperforming in drowsiness detection, the previous method had a critical drawback in generating representations. The previous method had a possibility that the method generates extremely sparse representation which cannot contain sufficient information to detect drowsiness. This work is improved and extended from our earlier work \cite{yu2016representation}, and we propose an end-to-end learning framework for a novel representation called condition-adaptive representation for drowsiness detection.

The condition-adaptive representation learning is a representation learning process to take the feature focused on some particular condition using auxiliary information (a.k.a., meta information). When the training dataset can be classified to several conditions, whilst the normal representation learning perform to extract generalized features from overall training data the condition-adaptive representation learning can extract more specific representations reflecting given conditions. Figure 1 represents the comparison of processes about the normal representation learning and condition-adaptive representation learning. An auxiliary information has been used to improve the performance of the deep learning model in many computer vision studies \cite{hong2016learning, zhang2016learning}. Hong et al. proposed deep learning system using transferrable knowledge to the scene segmentation in training phase \cite{hong2016learning}. Zhang et al. proposed a face alignment method using the result of landmark detection as auxiliary information \cite{zhang2016learning}. These methods tried to improve the performance of their solutions by learning the features biased to extra information that could help to explore useful features in their target domains. As with the methods described above, the concept of the condition-adaptive representation could be possibly interpreted as a representation biased to some conditions. However, in compared to the above methods which use extra information solely in training phase as prior knowledge, the proposed framework can generate the information which can help to improve the discrimination of the learnt representation during not only the training task but also testing task. By using this paradigm, the proposed framework can immediately generate the representation which adapts to the interpreted results.

The proposed framework is composed of four models consisting of representation learning, scene understanding, feature fusion, and drowsiness detection. The representation learning model discovers the rich and discriminative representation that can describe the motion and appearance of an object within the consecutive frames simultaneously. The scene understanding model identifies the various scene conditions that relate to driving conditions, e.g., illumination conditions and wearing glasses. The feature fusion model generates a condition-adaptive representation which is biased to a specific scene condition as opposed to the general spatio-temporal representation. The proposed framework detects drivers drowsiness in various situations accurately by using this condition-adaptive representation. The main contribution of this work is the representation learning framework that could be adapted to the particular scene conditions via understanding the scenes and generating the condition adaptive representation.

The rest of the paper is organized as follows. In Section \rom{2}, we give an overview of the 2D and 3D CNNs. The architectural detail of the proposed framework is explained in Section \rom{3}. We describe the training and inferencing procedure of the proposed framework in Section \rom{4}, and represent the method of data argumentation in Section \rom{5}. In Section \rom{6}, we show the experimental results and analysis those results. The conclusion and discussion are described in Section \rom{7}.


\begin{figure}
	
	\centering
	
	\includegraphics[width=\columnwidth]{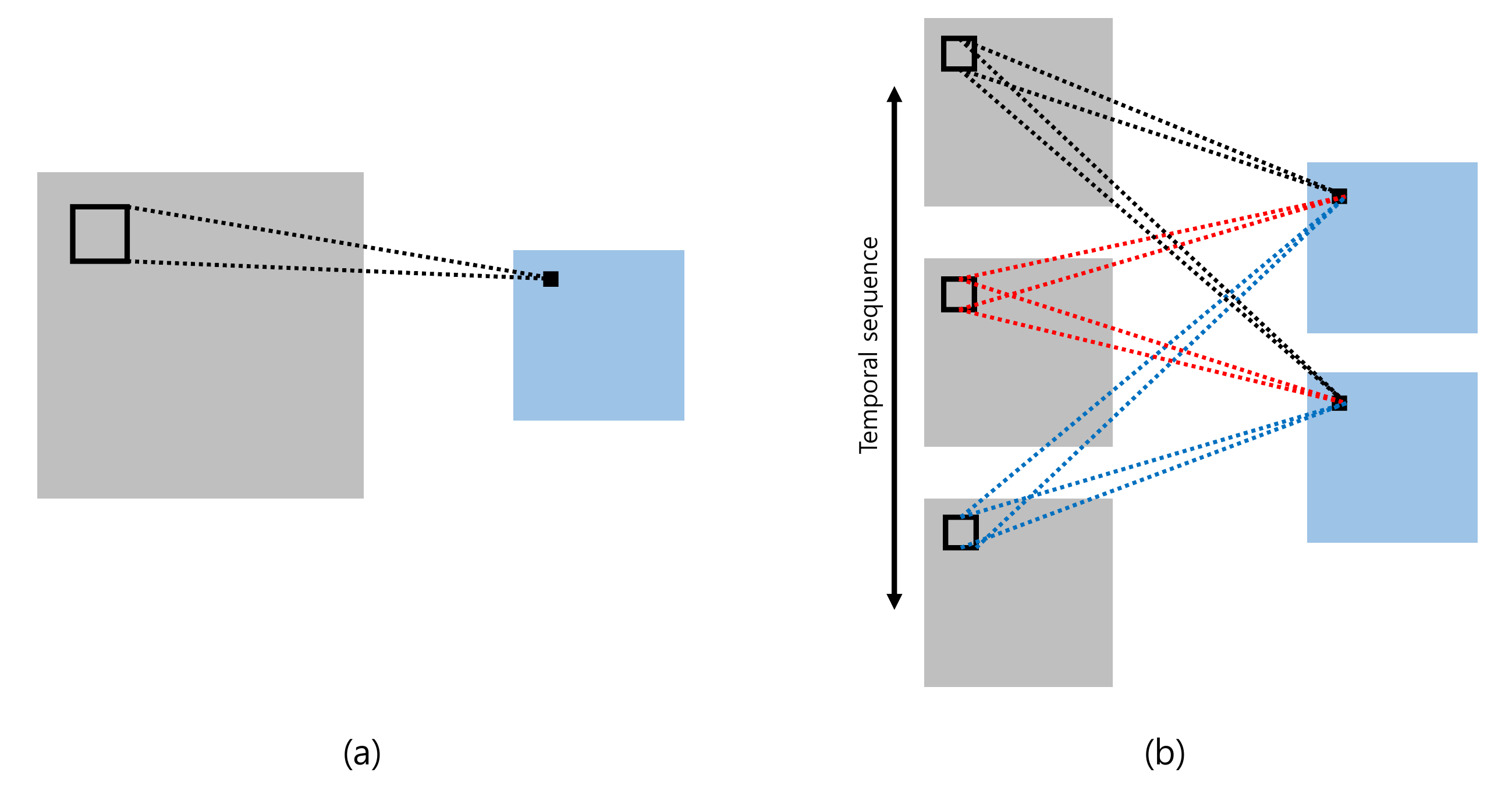}
	
	\caption{Illustrations of (a) 2D and (b) 3D convolution kernels. The connections sharing the same color denote a weight sharing in convolution layer. In 3D convolution (b), a temporal dimension is 3.}
	
	\label{fig:example}
	
\end{figure}

\section{2D and 3D Convolutional neural networks}

A convolutional neural network (CNN) (a.k.a., Deep convolutional neural network)  is a multi-layer weighted filter model introduced by LeCun et al. \cite{lecun1998gradient}. CNNs show outstanding performance in many computer vision studies such as image classification \cite{krizhevsky2012imagenet}, object detection, and recognition \cite{ren2015faster}. The key architectural characteristics of CNNs are ensuring some degree of shift, scale, and distortion invariance: local receptive field, shared weight, and spatial or temporal sub-sampling \cite{lecun1998gradient}. The function of a locally connected neural network in CNNs permits that CNNs can extract locally meaningful features, and by using the weight sharing, CNNs can be used as a elementary feature detector for one part of an image, across the set of entire images.

\begin{figure*}[ht]
	
	\vspace{-0.2cm}
	
	\begin{center}
		
		\centerline{\includegraphics[width=\textwidth]{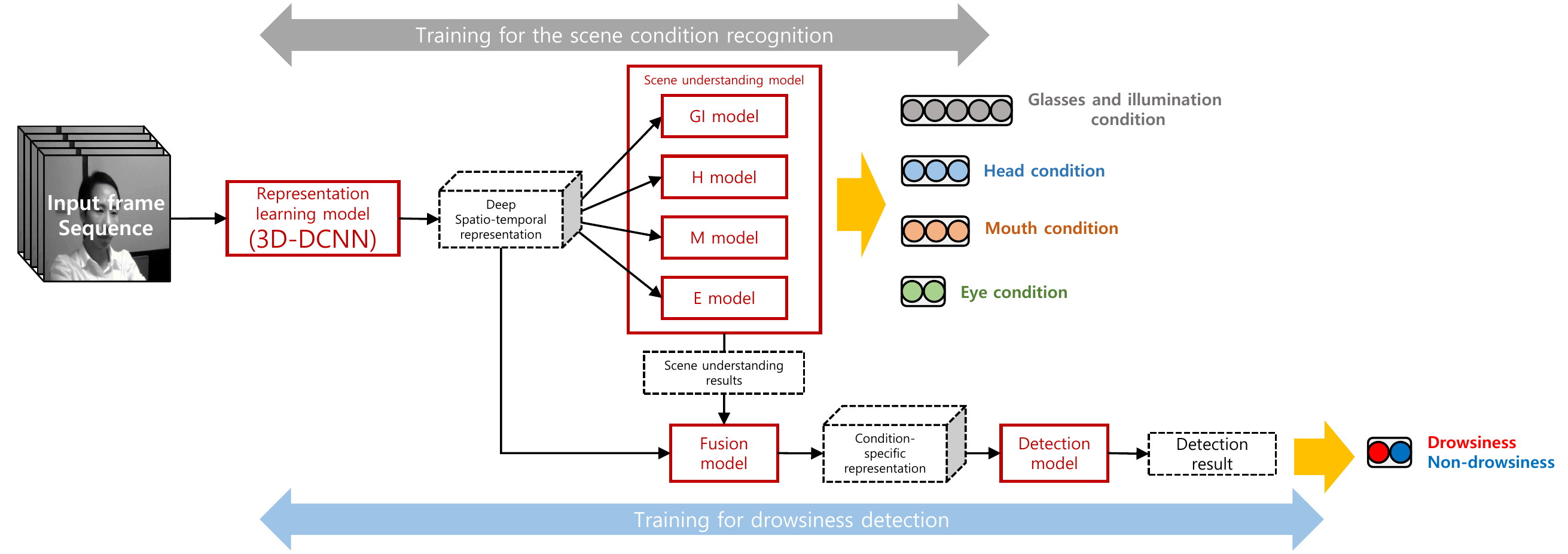}}
		
		\caption{Overall architecture of the proposed framework. The red boxes with bold line denote the models, and the black boxes drawn by dotted line define extracted features or outputs of each model.}
		
		\label{fig:example}
		
	\end{center}
	
	\vspace{-0.2cm}
	
\end{figure*}

In general CNNs, the convolution is performed at the convolution layers to discover features from spatial neighbourhoods on feature maps in each layer. Formally, the value of a unit at position $(x,y)$ in the $i$-th feature map in the $j$-th layer presented as $a^{xy}_{ij}$ is represented by

\begin{equation}
\begin{aligned}
a^{xy}_{ij}= \alpha[\sum^{W}_{p}\sum^{H}_{q}(v^{pq}w_{ij}^{pq})+b_{ij}]
\end{aligned}
\end{equation}

where $\alpha$ is the activation function such as hyperbolic tangent, sigmoid, and rectified linear functions, and $b_{ij}$ is the bias for the feature map, and $v$ is latent representation of the unit at position $(x,y)$ in the $i$-th feature map in the $j$-th layer. $w$ is the value of the kernel (Local receptive field) connected to the feature map, and  $W$ and $H$ are the width and the height of the kernel respectively. In the sub-sampling layer, the dimensional scale of the feature map is reduced by pooling over the spatially adjacent neighbourhood on the feature maps in the previous layer. The learnt feature using 2 dimensional-CNN (2D-CNN) can not only discover the locally useful feature but also be helpful to understand an entire image.

However, although the spatial features extracted from the 2D-CNN is robust to various computer vision studies, this paradigm of 2D-CNNs plays the role of a hurdle in learning the temporal representations about the sequential data such as video. To discover the rich and informative information from the sequential data using CNNs, Ji et al. proposed the 3D convolution \cite{ji20133d}. The 3D convolution is achieved by convolving a 3D feature map to the 3D volume formed by stacking multiple images together. By this principle, the feature maps in the convolution layers can capture temporal information that is contained in multiple contiguous frames. The value of a unit at position $(x,y,t )$ in the $i$-th feature map in the $j$-th layers which is denoted as $a^{xyt}_{ij}$ can be formulated as

\begin{equation}
\begin{aligned}
a^{xyt}_{ij}= \alpha[\sum^{W}_{p}\sum^{H}_{q}\sum^{D}_{k}(v^{pqk}w_{ij}^{pqk})+b_{ij}]
\end{aligned}
\end{equation}

where $\alpha$ is the activation function in 3D convolution, $v$ is latent representation of a unit at position $(x,y,t)$ in the $i$-th feature map in the $j$-th layer, $b_{ij}$ is the bias for the feature map, $w$ is the value of the kernel (3D Local receptive field) connected to the feature map, and $W$, $H$ and $D$ are the width, the height, and the depth of the kernel, respectively. Figure 2 shows the comparison of 2D and 3D convolutions. While the 2D convolution extracts spatial representation from given single image only, a 3D convolution can extract both spatial and temporal representation simultaneously in multiple consecutive images because the kernel of 3D convolution explore not only spatial axis but also temporal axis.

\section{Architecture}

\label{sec:blind}

The proposed framework is based on four models for representation learning, the scene understanding, the feature fusion, and the drowsiness detection. The representation learning model $f_{d}$ based on 3D-DCNN is used to extract the spatio-temporal representation from an input data. The scene understanding model consists of four sub-models $f_{gl}, f_{h}, f_{m}, f_{e}$ for interpreting the condition of glasses, illuminations, and movement of facial elements. The fusion model $f_{fu}$ generates condition-adaptive representation which can acclimatize the scene conditions. The detection model $f_{det}$ determines whether a driver is sleepy or not. Figure 3 shows an overall architecture of the proposed framework. The brief explanation for how to generate condition-adaptive representation and detect drowsiness of drivers, using the proposed framework is as follows. Initially, the representation learning based on the 3D-DCNN extracts a feature that can describe motion and appearance from a video clip simultaneously. Secondly, the scene understanding predicts five scene conditions that associated with wearing glasses, illumination conditions, and facial elements using the spatio-temporal feature extracted from the representation learning. The scene understanding results are represented by a vector that is defined by the one-hot encoding method. The one-hot encoding is one of the encoding approaches which indicates the state of a system using the binary values. The encoding result is represented by the group of bits among which the legal combinations of values are only those with a single high (1) bit and all the others low (0) bits. Then, feature fusion learns a condition-adaptive representation by agglomerating the spatio-temporal representation and the one-hot vectors. Finally, the detection model identifies a state of driver drowsiness by analyzing the condition-adaptive representation. In the following, we will describe the detail of information of each model and training scheme of the proposed framework.

\begin{figure*}[ht]
	
	\vspace{-0.2cm}
	
	\begin{center}
		
		\centerline{\includegraphics[width=\textwidth]{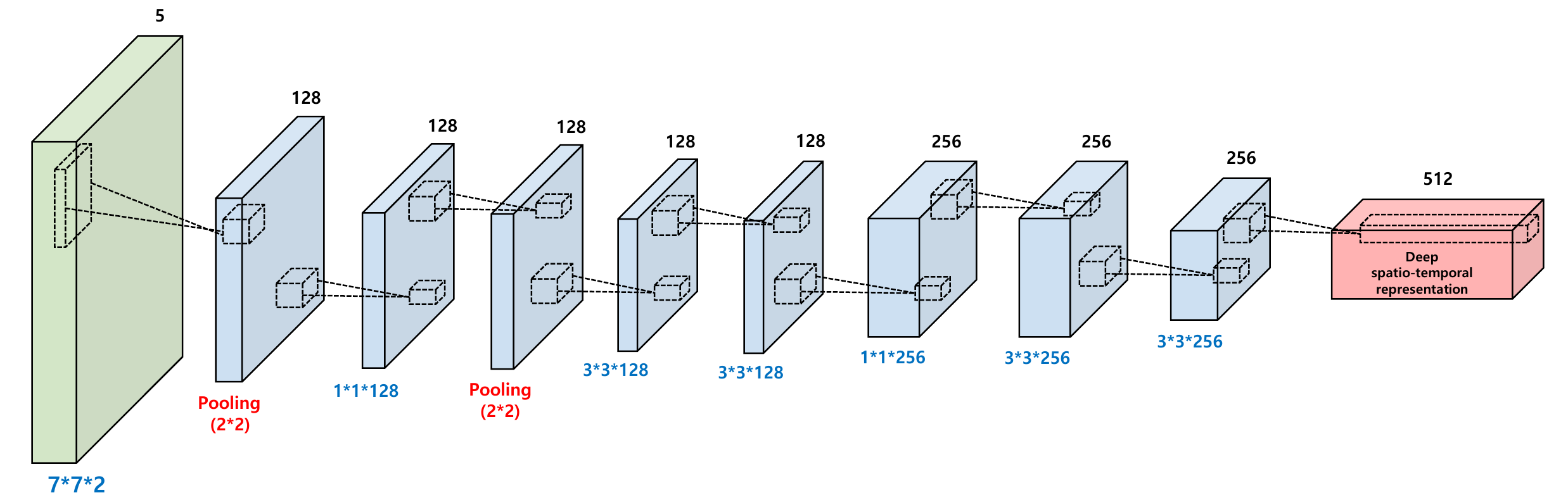}}
		
		\caption{Illustration of the 3D-DCNN in representation learning module. The green box and red box denote an input data and extracted spatio-temporal representation respectively, and the blue boxes represent convolution layers and pooling layers. Numbers located in the upside of the boxes represent the depth of each layer, and numbers below the boxes illustrate the dimensionality and structural detail of the kernel in each convolutional layer.}
		
		\label{fig:example}
		
	\end{center}
	
	\vspace{-0.2cm}
	
\end{figure*}

\subsection{Spatio-temporal representation learning}

In this section, we describe the representation learning model using 3D-DCNN for extracting the spatio-temporal representation from given mutlitple consecutive frames. The objective of the representation learning is discovering a rich and discriminative feature from inputted consecutive frames. Videos taken by the frontal facing camera in the display units of a vehicle can be variously modified depending on the various conditions of the vehicle interiors or exteriors, such as illumination conditions and an interior design of a vehicle. When drivers feel drowsiness, their facial elements make various changes, and these changes would be interpreted as either a shift in shape or change of motion. Therefore, to detect a drowsiness of drivers, we have to consider the representation which can describe spatial information (appearance) and temporal information (motion) simultaneously. It is impossible to estimate a temporal information using only a single frame since a single frame cannot contain a change according to a time sequence. When we consider these limitations observed when a input is a single frame, it is necessary to use multiple consecutive frames as an input to discover the spatial and temporal information simultaneously. In this work, we employed 3D-DCNN to discover various spatial and temporal change in given multiple consecutive frames.

Let $x \in R^{W \times H \times T}$ denotes a training video clip where $W$, $H$, and $T$ are the width, height, and the temporal length respectively. For a given input video clip $x$, the representation learning based on the 3D-DCNN extract a spatio-temporal representation as

\begin{equation}
\begin{aligned}
\boldsymbol{a} = f_{d}(x;\theta_{d}),\hspace{5mm} \boldsymbol{a} \in R^{W_{a}\times H_{a} \times D_{a}}
\end{aligned}
\end{equation}

where $\theta_{d}$ is the parameter vector of the representation learning, and $\boldsymbol{a}$ is a learnt spatio-temporal representation. The spatio-temporal representation is defined as the activation values of the hidden units in the last convolutional layer of 3D-DCNN of the representation learning model. $W_{a}$, $H_{a}$, and $D_{a}$ denote the width, height, and depth of the spatio-temporal representation. The 3D-DCNN in the representation learning is composed of six convolutional layers and two pooling layers. Figure 4 shows the architectural detail of the 3D-DCNN in the representation learning. To discover a spatial and temporal feature simultaneously, we employed a 3D local receptive field suggested by Tran et al. \cite{tran2015learning}. The convolutional operation based on 3D local receptive field can be defined as

\begin{equation}
\begin{aligned}
a= \rho[\sum^{W_r}_{i}\sum^{H_r}_{j}\sum^{D_r}_{k}(v_{i,j,k}w_{i,j,k}+b)]
\end{aligned}
\end{equation}

where $a$ is an activation value of the hidden unit, and $v$, $w$, and $b$ are the input value, the weight, and bias respectively. $W_{r}$, $H_{r}$, and $D_{r}$ denote the  width, the height, and the depth of 3D local receptive field, and $\rho$ is an activation function for the convolution layer. We adopt the Rectified Linear Units (ReLUs) \cite{krizhevsky2012imagenet} for the proposed 3D-DCNN. While the ordinary 2D structure of the kernel (local receptive field) in 2D convolution layers can extract spatial information only, the 3D structure of the kernel in 3D convolution layer allows to us capturing the spatial and temporal features simultaneously. The extracted representations which contain spatial and temporal features convey to the scene understanding model and feature fusion model to identify the various scene conditions and generate the condition-adaptive representation.

\subsection{Scene understanding}

The goal of the scene understanding is interpreting of the scenes with drivers, and understanding the various condition of drivers that can be categorized by the physiological and environmental conditions such as movement of facial elements, wearing glasses, and a difference between a day and night. These interpreted information help to train the framework for adapting the learnt representation to the various scene conditions. We hypothesize that each video clip is associated with the scene conditions and a driver drowsiness status. These are represented by either ground-truth (in training phase) or prediction results (in inferencing phase).

In this work, the scene condition contains the three categories of the facial elements and one category for the status of glasses and illumination: 1) conditions of glasses and illumination  $\mathcal{L}_{gl}$, 2) head $\mathcal{L}_{h}$, 3) mouth $\mathcal{L}_{m}$, and 4) eye $\mathcal{L}_{e}$. We define states of facial elements and the conditions for glasses wearing and illumination using a one-hot vector. The detailed explanation for the annotation of each scene condition is described in Table \rom{1}. We adopt a fully connected neural network since there is a possibility that given spatiotemporal representations have complex distributions which can not be modelled by a linear kernel. The predictions of conditions using the scene understanding model are written by   

\begin{equation}
\begin{aligned}
&\hat{\mathcal{L}_{gl}}=f_{gl}(\boldsymbol{a};\theta_{gl}),\hspace{5mm} \mathcal{L}_{gl} \in R^{L_{gl}\times 1}\\
&\hat{\mathcal{L}_{h}}=f_{h}(\boldsymbol{a};\theta_{h}),\hspace{5mm} \mathcal{L}_{h} \in R^{L_{h}\times 1}\\
&\hat{\mathcal{L}_{m}}=f_{m}(\boldsymbol{a};\theta_{m}),\hspace{5mm} \mathcal{L}_{m} \in R^{L_{m}\times 1}\\
&\hat{\mathcal{L}_{e}}=f_{e}(\boldsymbol{a};\theta_{e}),\hspace{5mm} \mathcal{L}_{e} \in R^{L_{e}\times 1}
\end{aligned}
\end{equation}

where $\hat{\mathcal{L}}\in\{\hat{\mathcal{L}_{gl}}, \hat{\mathcal{L}_{h}}, \hat{\mathcal{L}_{m}}, \hat{\mathcal{L}_{e}}\}$ are predicted scene conditions associated to input data $x$, and $L\in\{L_{gl}, L_{h}, L_{m}, L_{e}\}$ are dimensions of each annotation for the condition containing glasses and illumination, head, mouth, and eye. $\theta\in\{\theta_{gl},\theta_{h},\theta_{m},\theta_{e}\}$ are the parameters of the each model that defined by the fully connected network in the scene understanding model. Each model is composed of two hidden layers and a corresponding output layer. The aforementioned models are represented as

\begin{equation}
\begin{aligned}
o = f_{o}\{f_{h2}[f_{h1}(aW_{h1}+b_{h1})W_{h2}+b_{h2}]W_{o}+b_{o}\}\\
\end{aligned}
\end{equation}

where $f_{h1}$, $f_{h2}$, and $f_{o}$ are activation functions of the first and second hidden layers and an output layer respectively. $a$ is reshaped a spatio-temporal representation which is extracted from the representation learning model based on 3D-DCNN. $W_{h1}$, $W_{h2}$, and $W_{o}$ are weight parameters of two hidden layers and the output layer. $b_{h1}$, $b_{h2}$, and $b_{o}$ are the bias parameters of each layer. The learning procedure of each sub-model in the scene understanding is similar to the back propagation algorithm \cite{le1990handwritten}. Each sub-model estimates a condition that corresponding to the given spatio-temporal representations $\boldsymbol{a}$, then computes the difference between the predicted conditions and annotations to train the parameters of the network of the sub-model. The dimensionalities of the outputs for each scene understanding model correspond to their target domain to predict. For example, the dimensonality of the output $o$ of the scene understanding model for glasss and illumination conditions is five, because of the model is designed to identify the conditions defined as five classes. For a given spatio-temporal representation as input, the scene understanding model is trained to optimize the objective function defined as follows


\begin{equation}
\begin{aligned}
E_{su}(\hat{\mathcal{L}},\mathcal{L};\theta)=\min_{\theta_{d},\theta_{gl},\theta_{h},\theta_{m},\theta_{e}}\beta\sum_{i}[E_{gl}(\mathcal{L}_{gl}, \hat{\mathcal{L}}_{gl})\\+E_{h}(\mathcal{L}_h, \hat{\mathcal{L}}_h)+E_{m}(\mathcal{L}_m, \hat{\mathcal{L}}_m)+E_{e}(\mathcal{L}_e, \hat{\mathcal{L}}_e)].
\end{aligned}
\end{equation}

where $\mathcal{L}\in\{\mathcal{L}_{gl},\mathcal{L}_{h},\mathcal{L}_{m},\mathcal{L}_{e}\}$ denote annotations of input data, and $E_{gl},E_{h},E_{m}$, and $E_{e}$ denote loss functions defined by the softmax cross-entropy loss between the annotation and predicted results. $\beta$ is a hyper-parameter for regularization of the summation of values of error functions. The details of training and inference tasks are given in Section \rom{4}. The spatio-temporal representation and the outputs of the scene understanding model are then combined to produce the condition-adaptive representation explained in the following subsections.


\setlength{\tabcolsep}{2pt}

\begin{table}
	
	\begin{center}
		
		\caption{Annotations for the sub-models in the scene understanding and its status.}
		
		\label{table:headings}
		
		\begin{tabular}{ lclclclclcl } 
			
			\hline
			
			Scene condition & Category &One-hot vector & Condition \\
			
			\hline
			
			\multirow{5}{11em}{Glasses and illumination conditions} &1 &10000 & Day bare face \\ 
			
			&2& 01000 & Day glasses \\ 
			
			&3& 00100 & Night glasses \\ 
			
			&4& 00010 & Night bare face \\ 
			
			&5& 00001 & Day sunglasses \\ 
			
			\hline
			
			\multirow{3}{11em}{Head condition} &1& 100 & Normal status \\ 
			
			&2& 010 & Looking at both sides \\ 
			
			&3& 001 & Nodding \\ 
			
			\hline
			
			\multirow{3}{11em}{Mouth condition} &1& 100 & Normal status \\ 
			
			&2& 010 & Talking and laughing \\ 
			
			&3& 001 & Yawning \\ 
			
			\hline
			
			\multirow{2}{11em}{Eye condition} &1& 10 & Sleepiness eye \\ 
			
			&2& 01 & Normal status \\ 
			
			\hline
			
	\end{tabular}\end{center}
	
\end{table}

\setlength{\tabcolsep}{1.4pt}

\begin{figure*}[ht]
	
	\vspace{-0.2cm}
	
	\begin{center}
		
		\centerline{\includegraphics[width=\textwidth]{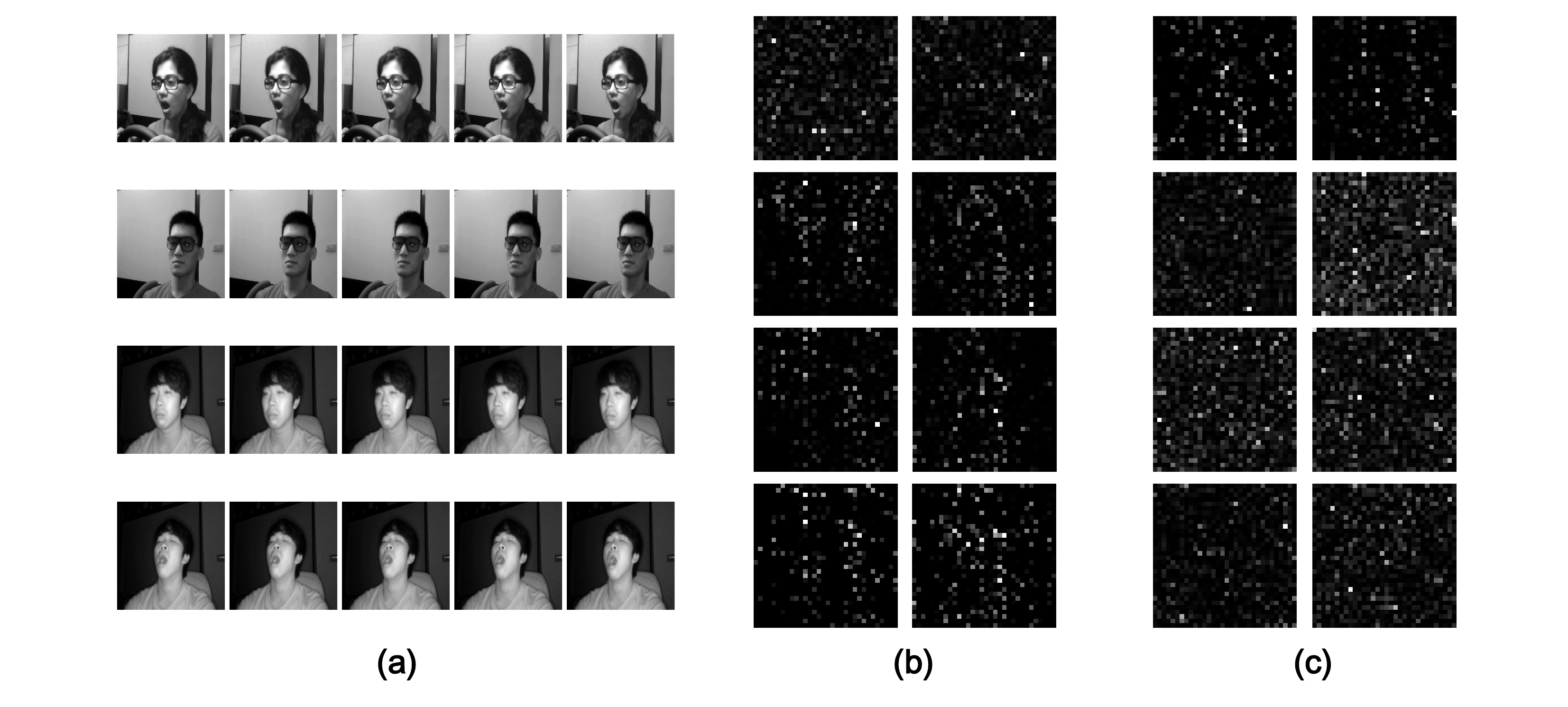}}
		
		\caption{Illustration of the deep spatio-temporal representation and condition-adaptive representation according to input data. (a) Input frames, (b) Deep spatio-temporal representation, and (c) denotes condition-adaptive representation obtained by the fusion model $f_{fu}$. Two images in (b) and (c) represents the visualization of activation results of hidden units in representation learning and feature fusion modules. The proposed condition-adaptive representation learning framework adaptively discover the conditional feature in an input volumes depending on the result of the scene understanding model.}\label{fig:example}
		
	\end{center}
	
	\vspace{-0.2cm}
	
\end{figure*}

\subsection{Feature fusion}

The objective of the model for feature fusion is to learn a set of condition-adaptive representations from the given spatio-temporal representation $\boldsymbol{\alpha}$ and its associated scene condition annotations  $\hat{\mathcal{L}}\in\{\hat{\mathcal{L}_{gl}}. \hat{\mathcal{L}_{h}}. \hat{\mathcal{L}_{m}}. \hat{\mathcal{L}_{e}}\}$.  Given the spatio-temporal representation extracted from 3D-DCNN $\boldsymbol{\alpha} \in R^{W_{\alpha} \times H_{\alpha} \times D_{\alpha}}$ and its associated and predicted scene conditions  $\hat{\mathcal{L}}$, the fusion model discovers a set of condition-adaptive representation $\beta$. The condition-adaptive feature vector $\beta$ is generated by using the multiplicative interaction approach proposed by Memisevic et al., \cite{memisevic2013learning}. Hong et al. observed that the high-order dependency between relevant features can be captured by using element-wise multiplication interaction between the feature maps \cite{hong2015learning}. To train the proposed framework that generates the combined representation which needs joint learning between the multiple resources, we refer to the training procedure proposed by Hong et al., \cite{hong2015learning}. The fusion model is defined as follows


\begin{equation}
\begin{aligned}
\beta=f_{fu}(\boldsymbol{\boldsymbol{\alpha}},\mathcal{L};\theta_{fu})
\end{aligned}
\end{equation}

\begin{equation}
\begin{aligned}
\beta =W_{fu}(W_{fea}\boldsymbol{\alpha} \otimes W_{gl}\mathcal{L}_{gl}\otimes W_{h}\mathcal{L}_{h}\\\otimes W_{m}\mathcal{L}_{m}\otimes W_{e}\mathcal{L}_{e})+b_{fu}.
\end{aligned}
\end{equation}

where $\beta$ denotes the unnormalized condition-adaptive representation, $b_{fu} \in R^{d\times 1}$ is the bias of the fusion model, and $\otimes$ denotes element-wise multiplication. The weights are given by $W_{fu} \in R^{M \times d}$, $W_{fea} \in R^{d \times W_{\alpha}H_{\alpha}D_{\alpha}}$, and $W_{gl}$, $W_{h}$, $W_{m}$, and $W_{e}$ are defined as the specific sizes based on the dimensional scale of each associated annotation. The variables $M$ and $d$ denote the number of hidden units in the fusion model. This 5-way tensor product can capture the correlation between the input domains containing the spatio-temporal representation and the scene conditions.

However, the element-wise multiplication with the spatio-temporal representation and the outputs of the scene understanding empirically computes values that are close to zero. These computed values can influence not only the result of the fusion model but also computational procedure when the multiplication results exceeded the range that can be represented by computation machine. We adopted a normalization scheme to prevent values close to zero for avoiding the computational errors and finding high-order dependency between the spatio-temporal representation and the identified scene conditions. To prevent computational error and to pay attention to only a scene condition, we normalize $\beta$ to $v$ using the softmax function in \cite{qi2016dynamic, xu2015show}. The normalization is formulated as follows

\begin{equation}
\begin{aligned}
v_i=\frac{exp(\beta_i)}{\sum_{j}exp(\beta_j)}
\end{aligned}
\end{equation}

where $\beta_i$ represents $i$-th element of the unnormalized joint feature, and $v_i$ is $i$-th element of the normalized fusion feature. Intuitively, $v$ represents a condition-adaptive representation defined over all spatio-temporal representations and the corresponding scene conditions. Figure 5 shows the input images, the spatio-temporal representations, and the condition-adaptive representations. The condition-adaptive representations are then used as an inputs to the detection model, which is explained in next section.

\subsection{Drowsiness detection}

The fusion model described in the previous subsection generates a set of condition-adaptive representations $v$, which provide scene adaptive features containing information of facial elements and illumination of drivers. The drowsiness detection of the proposed framework using the given condition-adaptive representation $v$ in Eq. (10) is carried out via additional neural networks. As same as the scene understanding model, we put an additional fully connected deep neural network on top of the fusion model as follow:

\begin{equation}
\begin{aligned}
o_{det} = f_{det}(v;\theta_{det}).
\end{aligned}
\end{equation}

where $o_{det}$ denotes the output of the detection model, and $\theta_{det}$ is the model parameter. The output of the fully connected network is consists of two units: non-drowsiness unit and drowsiness unit, to classify the drowsiness of a driver. To compute the likelihood of the driver drowsiness, we apply the soft-max function $\frac{e^{x_i}}{\sum_{k=1}^{2}e^{x_{k}}}$ which reflects the drowsiness and non-drowsiness degrees of input. Using the soft-max function, we can detect the driver drowsiness in each input. A high value of the non-drowsiness unit signifies that a driver in the input frames is likely to be awake, and a high value of the drowsiness unit signified that the driver is falling asleep. An optimization scheme for both $f_{fu}$ and $f_{det}$ operates under the detection objective. Our detection model is trained to minimize the detection loss using detection annotation associated with fusion feature, and representation as follows:


\begin{equation}
\begin{aligned}
\min_{\theta_{f},\theta_{det}}\sum_{i}E_{det}(o_{det}, \hat{o}_{det})
\end{aligned}
\end{equation}

where $\hat{o}_{det}$ is a ground-truth value that corresponds to each input data $x$, and $E_{det}$ denotes the objective function of the detection model. We used the softmax cross-entropy function as the objective function for $E_{det}$. The objective function is worked to all models embedding into the proposed framework.

\section{Training and Inference}

The training of the proposed framework has two objectives including the scene understanding objective in Eq. (7) and the drowsiness detection objective in Eq. (12), and the harmony of those two objectives is essential for achieving a superb locally optimized solution. Combining Eq. (7) and (12), the overall objective function is defined by


\begin{equation}
\begin{aligned}
\min_{\theta_{d},\theta_{sc},\theta_{f},\theta_{D}}\sum_{i}((1-\lambda)E_{su}(\mathcal{L}_c, \hat{\mathcal{L}}_c)+\lambda E_{det}(o_D, \hat{o}_D)) 
\end{aligned}
\end{equation}

where $\lambda$ is a parameter for balancing during training two modules for the scene understanding and drowsiness detection. The objective function can optimize the four modules of the proposed framework simultaneously. However, when we begin the training, we do not train the all models of the proposed framework simultaneously. The overall architecture (see Fig 2.) shows that the proposed framework is sharing the output of the representation learning model, and also denotes that the representation learning and scene understanding models can considerably influence to the other models (feature fusion and drowsiness detection). First, we train the representation learning and scene understanding models during $n$ steps. After that, we train all models containing the feature fusion and detection models.

To detect the drowsiness of drivers from input video clip, the proposed framework generates spatio-temporal representations using the representation learning, and then the spatio-temporal representation is used to understand scene conditions. these two pieces of information are combined to produce the condition-adaptive representation. Drowsiness is detected by using this condition-adaptive representation.

\section{Data augmentation}

The most general approach to reduce overfitting on a given training dataset is artificially enlarging the dataset using label-preserving transformations \cite{krizhevsky2012imagenet}. In this work, we apply the data augmentation based on horizontal transformation and image pyramid technique. This approach allows transformation of an image with very little computation so that we can make an additional dataset without huge computational load. We generate horizontally flipped images from the original images, and these original images and flipped images are transformed by using the image filtering methods based on the Gaussian filter. Figure 6 illustrates the procedure of the data augmentation. We conduct this by extracting training patches using various values of variations and training our proposed framework on this extended dataset. In our experiments, we used three different variations to generate additional training samples by using the image pyramid paradigm. These two types of data augmentation approaches can sufficiently increase the number of the training samples. Without this scheme, our proposed framework suffers from substantial overfitting, and it can converge to a poorly local optimized solution.

\section{Experiments}

\subsection{Benchmark dataset}

Previous studies \cite{khushaba2011driver, takei2005estimate, wakita2006driver} on driver drowsiness detection attempted to recognize small cases in the private dataset which is constructed in their own experimental environment for driver drowsiness detection. Abtahi et al. provided a publicly-available dataset for yawning detection \cite{abtahi2014yawdd}. However, it is still insufficient for a comprehensive drowsy driver study. We used the NTHU Drowsy Driver Dataset (NTHU-DDD Dataset) to demonstrate an efficiency of the proposed framework for the drivers drowsiness detection. It is too difficult and dangerous to construct a dataset for detecting of driver drowsiness detection in real situations. The NTHU-DDD dataset is composed of several videos containing a driver who was sitting on a car seat and playing a racing game with driving simulator wheel and pedals. The drivers in the dataset conducted various facial expressions during video recording. The total time of the entire dataset is about 9 and a half hours.

\begin{figure}
	
	\centering
	
	\includegraphics[width=\columnwidth]{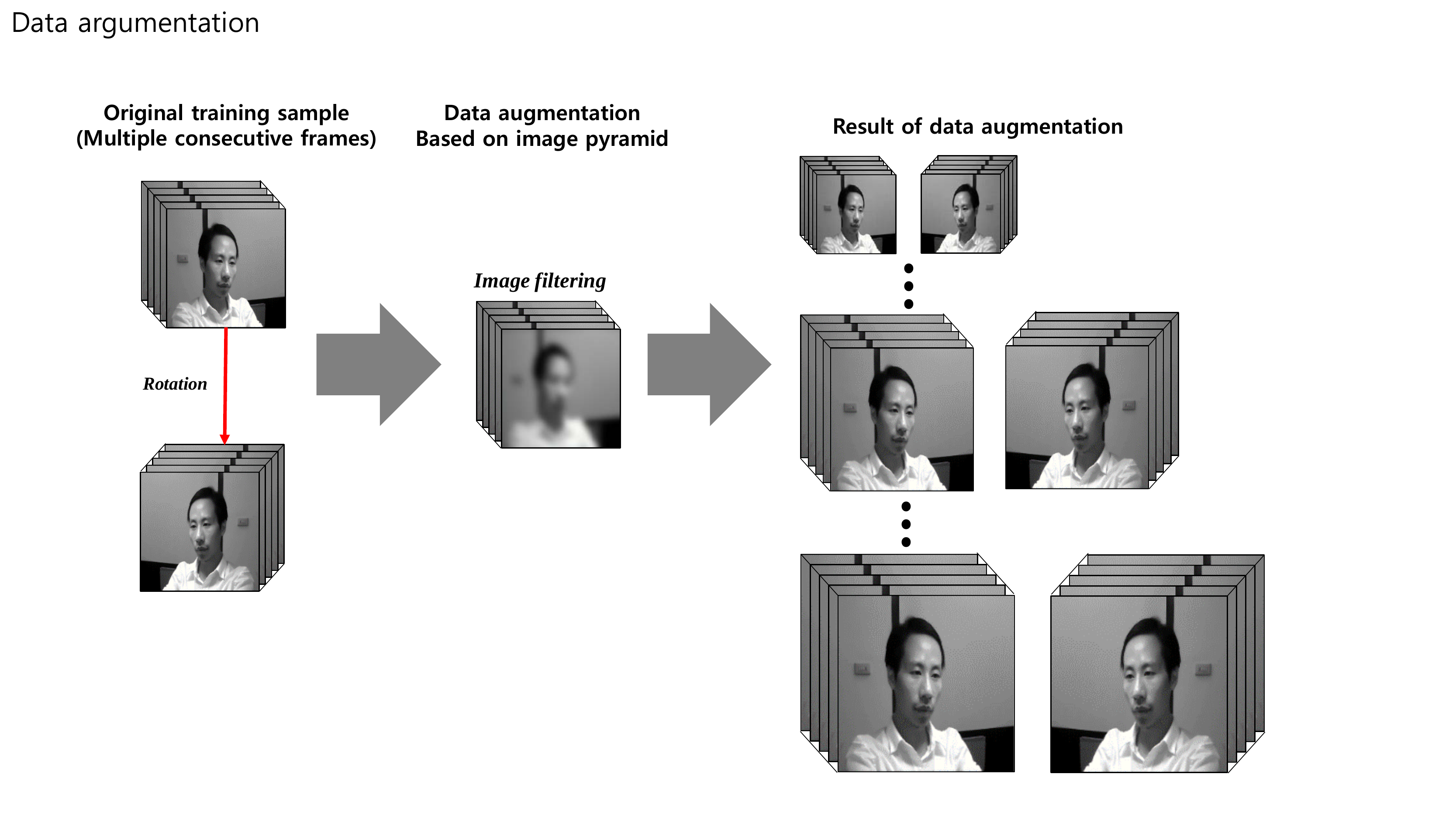}
	
	\caption{Illustration for the procedure of the data augmentation. Original training sample and the rotated sample of it generates another training samples by using the image filtering such as Gaussian filter.}
	
	\vspace{-0.2cm}
	
	\label{fig:example}
	
\end{figure}

The NTHU-DDD dataset is composed of three subsets for training, evaluation, and test, which are composed of non-redundant video files. Each subset consists of the videos which contain diverse situations for the condition for drivers that is captured using visual sensors such as a camera and an active infrared (IR) sensor. The entire dataset including training and evaluation datasets contain 36 of drivers of different ethnicities recorded with and without glasses/sunglasses under a variety of driving scenarios. The driving scenarios include normal driving, yawning, slow blink rate, falling asleep, and burst out laughing, under day and night illumination conditions. All videos contain frame-level annotation for the drowsiness condition. The video resolution is 640 $\times$ 480 in AVI format. Figure 7 shows example snapshots of the NTHU-DDD dataset.

\begin{figure*}[ht]
	
	\vspace{-0.2cm}
	
	\begin{center}
		
		\centerline{\includegraphics[width=\textwidth]{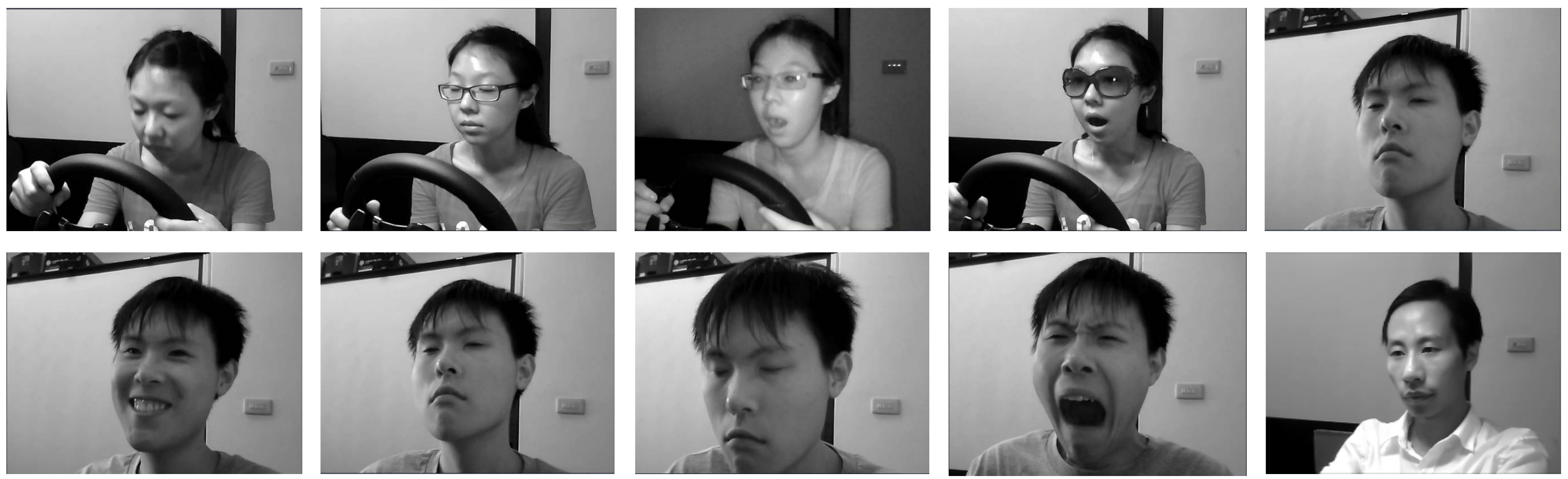}}
		
		\caption{The example snapshots of NTHU Drowsy Driver Detection Dataset (NTHU-DDD Dataset).}
		
		\label{fig:example}
		
	\end{center}
	
	\vspace{-0.2cm}
	
\end{figure*}

\begin{figure*}[ht]
	
	\vspace{-0.2cm}
	
	\begin{center}
		
		\includegraphics[width=\textwidth]{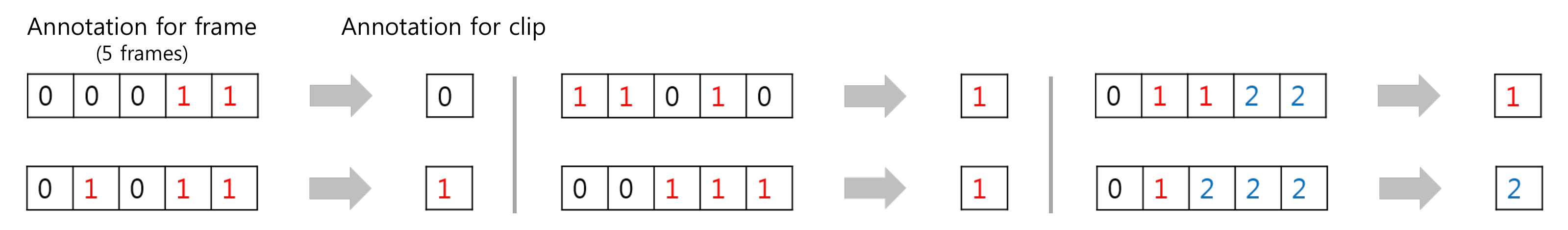}
		
		\caption{The illustration for the concept of temporal IOU.}
		
		\label{fig:example}
		
	\end{center}
	
	\vspace{-0.2cm}
	
\end{figure*}

The training dataset is composed of subsets that are composed of 18 subject folders. Each subject folder contains videos recorded in various driving condition. Each subset is classified into four scenarios defined as the condition of the glasses and illumination conditions (i.e., glasses, bare face, sunglasses, night glasses, night bare face). Each scenario contains four videos with different situation and corresponding annotation files. The evaluation dataset provides four subject folders and each subject contains five videos with different scenarios and corresponding annotation files. The training dataset is composed of 360 videos (722,223 frames), and the evaluation dataset contains 20 videos (173,259 frames). In this work, we only used training and evaluation datasets because test dataset can not publicly accessible and the test dataset not contains annotation for performance evaluation. We used all given training data to train the proposed framework. We make a small video clip that consists of five consecutive frames, and assign an annotation about the scene conditions and drowsiness status.


\setlength{\tabcolsep}{4pt}

\begin{table}
	
	\begin{center}
		
		\caption{Validation accuracies of the scene understanding model using the evaluation dataset in NTHU-DDD dataset.}
		
		\label{table:headings}
		
		\begin{tabular}{lclclclclcl}
			
			\hline\noalign{\smallskip}
			
			Scenario & Glasses and illumination & Head & Mouth & Eye \\
			
			\noalign{\smallskip}
			
			\hline
			
			\noalign{\smallskip}
			
			{Day bare face} & 0.99 & 0.99 & 0.98 & 0.89\\
			
			{Day glasses} & 0.97 & 0.93 & 0.95 & 0.81\\
			
			{Day sunglasses} & 0.98 & 0.97 & 0.78 & 0.78\\
			
			{Night bare face} & 0.99 & 0.95 & 0.97 & 0.82\\
			
			{Night glasses} & 0.97 & 0.96 & 0.88 & 0.92\\
			
			\hline
			
			{Average} & 0.98 & 0.96 & 0.912 & 0.844\\
			
			\hline
			
			{Total average} &  &  &  & 0.924\\
			
			\hline
			
		\end{tabular}
		
	\end{center}
	
\end{table}

\setlength{\tabcolsep}{1.4pt}

Unfortunately, the given training data provides frame-level annotation, so that we employed a concept of the intersection over union (IOU) \cite{farfade2015multi}, in order to change the frame-level annotation to clip-level annotation. Figure 8 shows the concept of the temporal IOU used in our experiment.  We assume that the annotation value of each clip is defined as a value occupying more than 50\% among the frame-level annotations. Therefore, we defined the annotation value as the value which is observed more than three frames in each clip in our experiment. In addition, we downsample all frames using a bilinear interpolation method in Opencv library to the uniform size with width of 224 pixels and height of 224 pixels for improving an experimental and time efficiencies.

\setlength{\tabcolsep}{4pt}

\begin{table*}[!t]
	
	\begin{center}
		
		\caption{Average accuracy comparison of the drowsiness detection approaches in different situations using the evaluation dataset in NTHU-DDD dataset. The \textbf{bolded values} represent the best accuracies in each scenario and the averages.}
		
		\label{table:headings}
		
		\begin{tabularx}{\textwidth}{lXlXlXlXlXlXlXlXlXl}
			
			\hline\noalign{\smallskip}
			
			Scenario & LeNet\cite{lecun1998gradient}  & AlexNet\cite{krizhevsky2012imagenet} & VGG-FaceNet\cite{parkhi2015deep} & LRCN\cite{donahue2015long} & FlowImageNet\cite{donahue2015long} & DDD-FFA\cite{parkdriver} & DDD-IAA\cite{parkdriver} &\textbf{Ours}\\
			
			\noalign{\smallskip}
			
			\hline
			
			\noalign{\smallskip}
			
			{Day bare face} & 0.531 & 0.704 &  0.638&  0.687& 0.563 & 0.782 & 0.698&  \textbf{0.796} \\
			
			{Day glasses}  & 0.592 & 0.616 & 0.705& 0.617& 0.616 & 0.741 & 0.759& \textbf{0.781} \\
			
			{Day sunglasses} & 0.682 & 0.702 & 0.570& 0.714& 0.675 & 0.618 & 0.698& \textbf{0.738} \\
			
			{Night bare face}& 0.602 & 0.646 & 0.737& 0.573& 0.668 & 0.702 & 0.749& \textbf{0.765} \\
			
			{Night glasses}& 0.599 & 0.627 & 0.741& 0.556& 0.551 & 0.683 & \textbf{0.747} & 0.734\\
			
			\hline
			
			{Average} & 0.601 & 0.659 & 0.678 & 0.629& 0.615 & 0.708 & 0.730& \textbf{0.762}\\
			
			\hline
			
		\end{tabularx}
		
	\end{center}
	
\end{table*}

\setlength{\tabcolsep}{1.4pt}

\setlength{\tabcolsep}{4pt}

\begin{table}
	
	\begin{center}
		
		\caption{F-measures and accuracies of the drowsiness detection using for the evaluation dataset in NTHU-DDD dataset. The listed values below the drowsiness and non-drowsiness attributes represent the results of F-measures.}
		
		\label{table:headings}
		
		\begin{tabular}{lclclclcl}
			
			\hline\noalign{\smallskip}
			
			Scenario & Drowsiness (F) & Non-drowsiness (F) & Accuracy\\
			
			\noalign{\smallskip}
			
			\hline
			
			\noalign{\smallskip}
			
			{Day bare face} & 0.809 & 0.784 & 0.796\\
			
			{Day glasses} & 0.789 & 0.774 & 0.781\\
			
			{Day sunglasses} & 0.758 & 0.718 & 0.738\\
			
			{Night bare face} & 0.753 & 0.777 & 0.765\\
			
			{Night glasses} & 0.718 & 0.750 & 0.734\\
			
			\hline
			
			{Average} & 0.765 & 0.760 & 0.762\\
			
			\hline
			
		\end{tabular}
		
	\end{center}
	
\end{table}

\setlength{\tabcolsep}{1.4pt}

\subsection{Experimental results}

We demonstrate an efficiency of our framework using the evaluation set of the NTHU-DDD dataset. The evaluation dataset is composed of 5 scenarios, and each scenario contains five videos that captured various virtual driving situations. The videos in the evaluation dataset are not duplicated to the videos in the training dataset. The dataset also includes multiple annotations that are concerned with the scene conditions and drowsiness detection. We tested the performances of the scene understanding and drowsiness detection respectively.

The scene understanding module is evaluated by using validation accuracy, represented as $\frac{n}{m}$ where the numerator $n$ is the number of the correctly classified results of each sub-model in the scene understanding model, and the denominator $m$ denotes the total number of test samples. Table \rom{2} shows the validation accuracies of the scene understanding model that is composed of four sub-models: the glasses and illumination conditions $f_{gl}$, the head model $f_{h}$, mouth model $f_{m}$, and eye model $f_{e}$. The averages are computed by the formulation of the arithmetic mean so that the weights according to the number of data that classified to the same categories in the table did not consider. This measurement has been applied equally to subsequent experiments. The average of validation accuracies across to all scene conditions for sub-models is 0.924. Experimental results in Table \rom{2} show that the scene understanding module in the proposed framework achieves good classification results in the classification problems of the glasses and illumination conditions and the status of a head. However, the classification result for the condition of mouth and eye is relatively lower than the other categories. The performance gaps between the sub-models in the scene understanding could be interpreted as a bias of representation learning. The understanding of the scene conditions based on our spatio-temporal representation could be influenced by the geometrical size and scale of a target object. Since the portion of each frame for an eye and mouth is relatively smaller than the portion of a frame for glasses, illumination, and head in the NTHU-DDD dataset, the learnt representation learning model would have been over-fitted to the conditions for glasses, illumination and head.

We evaluated the proposed framework quantitatively by using the F-measure. F-measure is harmonic mean of precision and detection rate, where precision and recall are defined as follows:

\begin{equation}
\begin{aligned}
Precision = \frac{TP}{TP+FP}
\end{aligned}
\end{equation}

\begin{equation}
\begin{aligned}
Detectionrate(DR) = \frac{TP}{TP+FN}
\end{aligned}
\end{equation}

\begin{equation}
	\begin{aligned}
	F-measure = \frac{2\times Precision \times DR}{Precision + DR}
	\end{aligned}
\end{equation}

where $TP$ (True positive) is the number of correctly detected as drowsiness state, and $FN$ (False negative) is the number of incorrect detection results that classified to non-drowsiness condition. $FP$ (False positive) is the number of non-drowsiness detection result incorrectly identified to the drowsiness state, and $TN$ (True negative) is the number of correctly classified as non-drowsiness state. The quantitative evaluation denotes an average over all videos represented as same glass and illumination categories. Table \rom{4} shows the accuracy of the proposed framework for the drowsiness detection. The results show that our proposed framework achieves an average accuracy of 0.762.

Due to the lack of performance comparison using a publicly available dataset for drowsiness detection, we referred the previous method which was evaluated their performance using the NTHU-DDD dataset or implement a method based on the well-known multiclass classification algorithm for images. We compared our framework to several methods \cite{parkhi2015deep, donahue2015long, parkdriver, krizhevsky2012imagenet, lecun1998gradient}. Parkhi et al. proposed a face recognition method (VGG-FaceNet) using a deep neural network \cite{parkhi2015deep}. The VGG-FaceNet consists of 36 convolution layers, and this network is much deeper than the 3D-DCNN used in the proposed framework. Donahue et al. provide the method based on long-term recurrent convolutional networks (LRCN) for visual recognition and description for long-term time series data \cite{donahue2015long}. We modified these methods to evaluate the performance of driver drowsiness detection. Park et al. proposed the deep drowsiness detection (DDD) network for drowsiness detection using feature-fused architecture \cite{parkdriver}. Park et al. used two different fusion strategies to their network: independently-averaged architecture (IAA) and feature-fused architecture(FFA). They provide the experimental results using the NTHU-DDD dataset. These methods were trained and tested with the equal procedure of the proposed framework. Additionally, we compare the results using the NTHU-DDD dataset, which is listed in Part et al.\cite{parkdriver}.

\begin{figure}
	
	\vspace{-0.2cm}
	
	\begin{center}
		
		\includegraphics[width=\columnwidth]{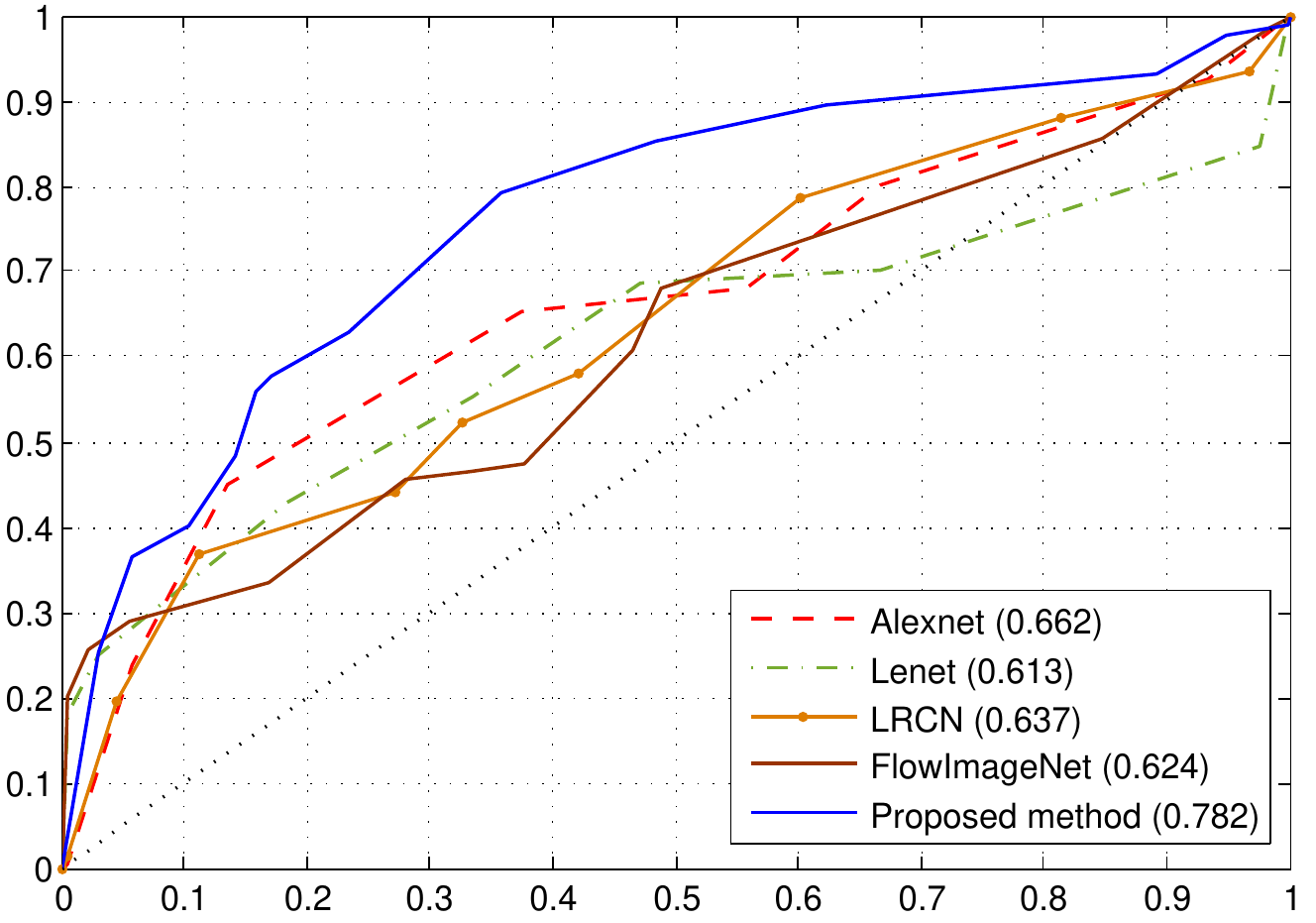}
		
		\caption{The ROCs for the driver drowsiness detection. Figures in parentheses indicate the area under curves (AUCs).}
		
		\label{fig:example}
		
	\end{center}
	
	\vspace{-0.2cm}
	
\end{figure}


\begin{figure*}[ht]
	
	\vspace{-0.2cm}
	
	\begin{center}
		
		\centerline{\includegraphics[width=\textwidth]{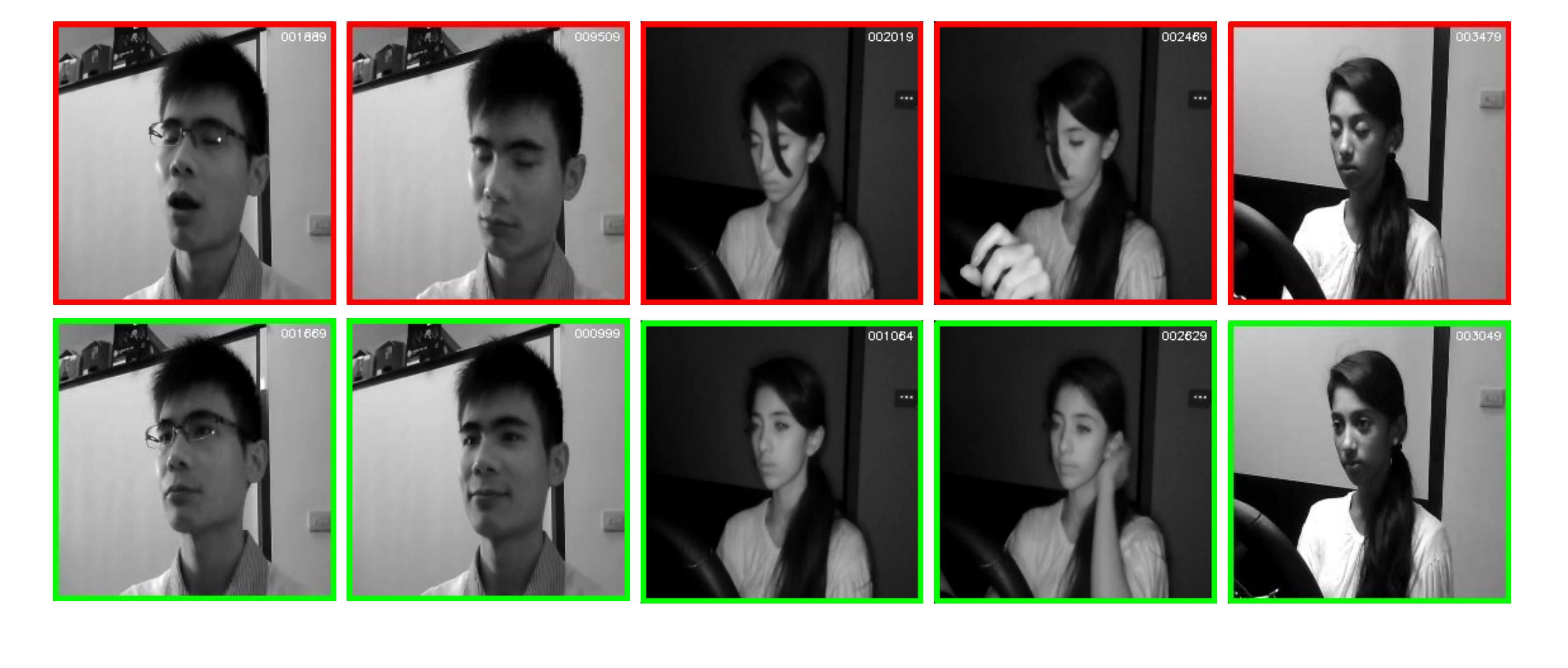}}
		
		\caption{The detection results using NTHU-DDD dataset. The images of the first row show the detection results for the driver drowsiness, and the images of the second row denote the detection results of a normal condition of drivers.}
		
		\label{fig:example}
		
	\end{center}
	
	\vspace{-0.2cm}
	
\end{figure*}

Table \rom{3} shows that the comparison results of driver drowsiness detection using NTHU-DDD dataset. The experimental results show that the proposed framework outperforms other methods in most of the scenarios. Only in the night glasses scenario did the proposed method achieve a performance lower than the DDD-IAA. Additionally, the experimental results illustrate that the proposed framework achieves higher and stable performance in various scene conditions than the listed methods, even though several methods used the deeper network structure. Figure 9 shows the receiver operating characteristic (ROC) curves and the area under curves (AUCs), generated by the evaluation dataset predictions. The results of the ROC plots in Fig. 9 present that the proposed method does not take a benefit in the lower regions of the curve, where the false positive rate (FPR) is less than 0.05 approximately, but provides a definite benefit for much of the rest of the curve, over the other methods \cite{lecun1998gradient, krizhevsky2012imagenet, parkhi2015deep, donahue2015long}.

The overall experimental results demonstrate that the proposed method can provide an accurate and effective method for the driver drowsiness detection than the other drowsiness detection method based on a visual analysis. Driver's drowsiness in the real world could appear with various variations of facial elements in diverse illumination conditions. The feature fusion helps to discover the discriminative and rich condition-adaptive representation for detecting the drowsiness, and this function plays a significant role to provide high-quality drowsiness detection in various situations. Figure 10  shows the example snapshots of the correct detection results using NTHU-DDD dataset.

\subsection{Computational complexity}

Although the computational cost of the framework depends on the size of input images and the structure details such as the number of layers and the size of kernels in a neural network, theoretically, the computational complexity of representation learning and feature fusion models based on 3D-CNN is $O\left(\sum_{i=1}^{d}W_{i} H_{i} D_{i} n_{i} m_{i} k_{i} \right)$, 

where $i$ and $d$ are the index of a convolutional layer and the number of convolutional layers of each model. $W_{i}$, $H_{i}$, and $D_{i}$ denote the width, height, and depth of input data in each convolutional layer. $n_{i}$, $m_{i}$, and $k_{i}$ denote the width, height, and depth of 3D-convolutional kernel in $i$-th layer. The computational complexity of the scene understanding and drowsiness detection models using two-layers neural networks is $O(N^2C)$, where $N$ and $C$ denote the dimensionalities of each hidden layer and target domain for objectives. We have estimated the computational complexity of the proposed framework based on the approaches of He et al., \cite{he2015convolutional} and Notchenko et al., \cite{notchenko2016sparse}.

Note these computational complexities apply to both training and testing phases, however practical execution times in both phases are different since the proposed framework shows different work-flows in training and test phases. The training task consists of the three steps: 1) calculation of output, 2) computing an error, and 3) updating the parameters. Therefore, the execution time in the training task is relatively longer than the time in the testing task. Once the model training end, the execution time in testing phase is much faster because of the framework only needs to compute the output for drowsiness detection. The execution time in our experimental setting was 38.1 FPS (28.6 $ms$) which is almost real-time, and was obtained. We calculated this value by averaging the execution time of the proposed framework for 300 seconds, except displaying an output on a screen.

The proposed framework is implemented with Google Tensorflow library. Although the training in the framework requires long times, after the model training is finished, the entire framework is able to perform in real-time with Python implementation using a Core i7, 3.4GHz PC with 16GB RAM and GTX TITAN GPU.

\section{Discussion and Conclusion}

In this paper, we have proposed an condition-adaptive representation learning for efficient driver drowsiness detection method which is invariant to various driving conditions containing a driving time such as day and night and a driver's appearance. To this end, we extracted the spatio-temporal representation and merged it with the vectors that represent the scene understanding results using the feature fusion method based on the tensor product approach. These problems are effectively modelled using 3D-DCNN and fully connected neural network based on recent advances in computer vision fields. The spatio-temporal representation and estimated scene conditions are merged to enhance the discriminative power for providing precise driver drowsiness detection in various driving conditions. With the feature fusion properly harnessed, the merged feature can provide more discrimination than the original spatio-temporal representation even though the original representation contains the motion and appearance information about the driving and drivers conditions. Experimental results show that the proposed framework outperforms other methods, including methods based on deep learning, in drowsiness detection accuracies.

The limitation of the proposed framework can be summarized as follows. First, although the proposed framework achieves good detection performance, it also needs a high-performance GPU computing unit that must be installed on a vehicle. It may cause high price of the vehicle and an increase in vehicle weight. Second, the proposed method needs many training samples that are labelled with the scene conditions and drowsiness state, for learning the representation that can cover various situations about drivers. Third, since the proposed framework is an off-line method, it can not guarantee to detect the drowsiness of drivers of entirely different types that are not included in training samples.

In future works, several suggestions should be taken into account. First, we will optimize the network structure in the proposed framework for use in an embedded board or microcomputing systems to reduce the financial cost and improve the computational efficiency without performance degradation. Second, we will develop an on-line updating method in order to improve the drowsiness detection reliability of the model through continuous updating. Third, we will study a data augmentation method based on generative models to improve the performance of drowsiness detection by enlarging the scale and variety of a given dataset.

\section*{Acknowledgment}
This work was supported by the Institute for Information and Communications Technology Promotion (IITP) grant funded by the Korea government (MSIP) (No. B0101-15-0525, Development of global multi-target tracking and event prediction techniques based on real-time large-scale video analysis), and the National Strategic Project-Fine particle of the National Research Foundation of Korea(NRF) funded by the Ministry of Science and ICT(MSIT), the Ministry of Environment(ME), and the Ministry of Health and Welfare(MOHW) (NRF-2017M3D8A1092022).



\ifCLASSOPTIONcaptionsoff

\newpage

\fi



\bibliographystyle{ieeetr}

\bibliography{egbibv2}

\begin{thebibliography}{10}

\bibitem{schroeder2013national}
P.~Schroeder, M.~Meyers, S.~Kostyniuk, Lidia, {\em et~al.}, ``National survey
  on distracted driving attitudes and behaviors-2012,'' Tech. Rep. DOT HS 811
  729, United States. National Highway Traffic Safety Administration,
  Washington, DC, 2013.

\bibitem{maycock1996sleepiness}
G.~Maycock, ``Sleepiness and driving: the experience of uk car drivers,'' {\em
  Journal of Sleep Research}, vol.~5, no.~4, pp.~229--231, 1996.

\bibitem{sagberg1999road}
F.~Sagberg, ``Road accidents caused by drivers falling asleep,'' {\em Accident
  Analysis \& Prevention}, vol.~31, no.~6, pp.~639--649, 1999.

\bibitem{pack1995characteristics}
A.~I. Pack, A.~M. Pack, E.~Rodgman, A.~Cucchiara, D.~F. Dinges, and C.~W.
  Schwab, ``Characteristics of crashes attributed to the driver having fallen
  asleep,'' {\em Accident Analysis \& Prevention}, vol.~27, no.~6,
  pp.~769--775, 1995.

\bibitem{garcia2012vision}
I.~Garcia, S.~Bronte, L.~M. Bergasa, J.~Almaz{\'a}n, and J.~Yebes,
  ``Vision-based drowsiness detector for real driving conditions,'' in {\em
  Intelligent Vehicles Symposium (IV), 2012 IEEE}, pp.~618--623, IEEE, 2012.

\bibitem{mbouna2013visual}
R.~O. Mbouna, S.~G. Kong, and M.-G. Chun, ``Visual analysis of eye state and
  head pose for driver alertness monitoring,'' {\em IEEE transactions on
  intelligent transportation systems}, vol.~14, no.~3, pp.~1462--1469, 2013.

\bibitem{wang2012method}
P.~Wang and L.~Shen, ``A method of detecting driver drowsiness state based on
  multi-features of face,'' in {\em Image and Signal Processing (CISP), 2012
  5th International Congress on}, pp.~1171--1175, IEEE, 2012.

\bibitem{minkov2012comparison}
K.~Minkov, S.~Zafeiriou, and M.~Pantic, ``A comparison of different features
  for automatic eye blinking detection with an application to analysis of
  deceptive behavior,'' in {\em Communications Control and Signal Processing
  (ISCCSP), 2012 5th International Symposium on}, pp.~1--4, IEEE, 2012.

\bibitem{panning2011color}
A.~Panning, A.~Al-Hamadi, and B.~Michaelis, ``A color based approach for eye
  blink detection in image sequences,'' in {\em Signal and Image Processing
  Applications (ICSIPA), 2011 IEEE International Conference on}, pp.~40--45,
  IEEE, 2011.

\bibitem{kurylyak2012detection}
Y.~Kurylyak, F.~Lamonaca, and G.~Mirabelli, ``Detection of the eye blinks for
  human's fatigue monitoring,'' in {\em Medical Measurements and Applications
  Proceedings (MeMeA), 2012 IEEE International Symposium on}, pp.~1--4, IEEE,
  2012.

\bibitem{suzuki2006measurement}
M.~Suzuki, N.~Yamamoto, O.~Yamamoto, T.~Nakano, and S.~Yamamoto, ``Measurement
  of driver's consciousness by image processing-a method for presuming driver's
  drowsiness by eye-blinks coping with individual differences,'' in {\em 2006
  IEEE International Conference on Systems, Man and Cybernetics}, vol.~4,
  pp.~2891--2896, IEEE, 2006.

\bibitem{khushaba2011driver}
R.~N. Khushaba, S.~Kodagoda, S.~Lal, and G.~Dissanayake, ``Driver drowsiness
  classification using fuzzy wavelet-packet-based feature-extraction
  algorithm,'' {\em IEEE Transactions on Biomedical Engineering}, vol.~58,
  no.~1, pp.~121--131, 2011.

\bibitem{patel2011applying}
M.~Patel, S.~Lal, D.~Kavanagh, and P.~Rossiter, ``Applying neural network
  analysis on heart rate variability data to assess driver fatigue,'' {\em
  Expert systems with Applications}, vol.~38, no.~6, pp.~7235--7242, 2011.

\bibitem{tran2010improving}
Y.~Tran, A.~Craig, N.~Wijesuriya, and H.~Nguyen, ``Improving classification
  rates for use in fatigue countermeasure devices using brain activity,'' in
  {\em 2010 Annual International Conference of the IEEE Engineering in Medicine
  and Biology}, pp.~4460--4463, IEEE, 2010.

\bibitem{papadelis2007monitoring}
C.~Papadelis, Z.~Chen, C.~Kourtidou-Papadeli, P.~D. Bamidis, I.~Chouvarda,
  E.~Bekiaris, and N.~Maglaveras, ``Monitoring sleepiness with on-board
  electrophysiological recordings for preventing sleep-deprived traffic
  accidents,'' {\em Clinical Neurophysiology}, vol.~118, no.~9, pp.~1906--1922,
  2007.

\bibitem{ersal2010model}
T.~Ersal, H.~J. Fuller, O.~Tsimhoni, J.~L. Stein, and H.~K. Fathy,
  ``Model-based analysis and classification of driver distraction under
  secondary tasks,'' {\em IEEE transactions on intelligent transportation
  systems}, vol.~11, no.~3, pp.~692--701, 2010.

\bibitem{yang2009detection}
J.~H. Yang, Z.-H. Mao, L.~Tijerina, T.~Pilutti, J.~F. Coughlin, and E.~Feron,
  ``Detection of driver fatigue caused by sleep deprivation,'' {\em IEEE
  Transactions on systems, man, and cybernetics-part A: Systems and humans},
  vol.~39, no.~4, pp.~694--705, 2009.

\bibitem{liu2009predicting}
C.~C. Liu, S.~G. Hosking, and M.~G. Lenn{\'e}, ``Predicting driver drowsiness
  using vehicle measures: Recent insights and future challenges,'' {\em Journal
  of safety research}, vol.~40, no.~4, pp.~239--245, 2009.

\bibitem{takei2005estimate}
Y.~Takei and Y.~Furukawa, ``Estimate of driver's fatigue through steering
  motion,'' in {\em 2005 IEEE international conference on systems, man and
  cybernetics}, vol.~2, pp.~1765--1770, Ieee, 2005.

\bibitem{wakita2006driver}
T.~Wakita, K.~Ozawa, C.~Miyajima, K.~Igarashi, I.~Katunobu, K.~Takeda, and
  F.~Itakura, ``Driver identification using driving behavior signals,'' {\em
  IEICE TRANSACTIONS on Information and Systems}, vol.~89, no.~3,
  pp.~1188--1194, 2006.

\bibitem{dinges1998perclos}
D.~F. Dinges and R.~Grace, ``Perclos: A valid psychophysiological measure of
  alertness as assessed by psychomotor vigilance,'' {\em US Department of
  Transportation, Federal Highway Administration}, no.~FHWA-MCRT-98-006, 1998.

\bibitem{dwivedi2014drowsy}
K.~Dwivedi, K.~Biswaranjan, and A.~Sethi, ``Drowsy driver detection using
  representation learning,'' in {\em Advance Computing Conference (IACC), 2014
  IEEE International}, pp.~995--999, IEEE, 2014.

\bibitem{dalal2005histograms}
N.~Dalal and B.~Triggs, ``Histograms of oriented gradients for human
  detection,'' in {\em Computer Vision and Pattern Recognition, 2005. CVPR
  2005. IEEE Computer Society Conference on}, vol.~1, pp.~886--893, IEEE, 2005.

\bibitem{lienhart2002extended}
R.~Lienhart and J.~Maydt, ``An extended set of haar-like features for rapid
  object detection,'' in {\em Image Processing. 2002. Proceedings. 2002
  International Conference on}, vol.~1, pp.~I--I, IEEE, 2002.

\bibitem{mcdonald2018contextual}
A.~D. McDonald, J.~D. Lee, C.~Schwarz, and T.~L. Brown, ``A contextual and
  temporal algorithm for driver drowsiness detection,'' {\em Accident Analysis
  \& Prevention}, vol.~113, pp.~25--37, 2018.

\bibitem{sayed2001unobtrusive}
R.~Sayed and A.~Eskandarian, ``Unobtrusive drowsiness detection by neural
  network learning of driver steering,'' {\em Proceedings of the Institution of
  Mechanical Engineers, Part D: Journal of Automobile Engineering}, vol.~215,
  no.~9, pp.~969--975, 2001.

\bibitem{krajewski2009detecting}
J.~Krajewski, M.~Golz, and D.~Sommer, ``Detecting sleepy drivers by pattern
  recognition based analysis of steering wheel behaviour,'' {\em Der Mensch im
  Mittelpunkt technischer Systeme}, pp.~288--291, 2009.

\bibitem{hearst1998support}
M.~A. Hearst, S.~T. Dumais, E.~Osuna, J.~Platt, and B.~Scholkopf, ``Support
  vector machines,'' {\em IEEE Intelligent Systems and their applications},
  vol.~13, no.~4, pp.~18--28, 1998.

\bibitem{simonyan2014very}
K.~Simonyan and A.~Zisserman, ``Very deep convolutional networks for
  large-scale image recognition,'' {\em arXiv preprint arXiv:1409.1556}, 2014.

\bibitem{girshick2016region}
R.~Girshick, J.~Donahue, T.~Darrell, and J.~Malik, ``Region-based convolutional
  networks for accurate object detection and segmentation,'' {\em IEEE
  transactions on pattern analysis and machine intelligence}, vol.~38, no.~1,
  pp.~142--158, 2016.

\bibitem{erhan2014scalable}
D.~Erhan, C.~Szegedy, A.~Toshev, and D.~Anguelov, ``Scalable object detection
  using deep neural networks,'' in {\em Proceedings of the IEEE Conference on
  Computer Vision and Pattern Recognition}, pp.~2147--2154, 2014.

\bibitem{ren2015faster}
S.~Ren, K.~He, R.~Girshick, and J.~Sun, ``Faster r-cnn: Towards real-time
  object detection with region proposal networks,'' in {\em Advances in neural
  information processing systems}, pp.~91--99, 2015.

\bibitem{molchanov2016online}
P.~Molchanov, X.~Yang, S.~Gupta, K.~Kim, S.~Tyree, and J.~Kautz, ``Online
  detection and classification of dynamic hand gestures with recurrent 3d
  convolutional neural network,'' in {\em Proceedings of the IEEE Conference on
  Computer Vision and Pattern Recognition}, pp.~4207--4215, 2016.

\bibitem{qi2016dynamic}
X.~Qi, C.-G. Li, G.~Zhao, X.~Hong, and M.~Pietik{\"a}inen, ``Dynamic texture
  and scene classification by transferring deep image features,'' {\em
  Neurocomputing}, vol.~171, pp.~1230--1241, 2016.

\bibitem{simonyan2014two}
K.~Simonyan and A.~Zisserman, ``Two-stream convolutional networks for action
  recognition in videos,'' in {\em Advances in neural information processing
  systems}, pp.~568--576, 2014.

\bibitem{du2015hierarchical}
Y.~Du, W.~Wang, and L.~Wang, ``Hierarchical recurrent neural network for
  skeleton based action recognition,'' in {\em Proceedings of the IEEE
  Conference on Computer Vision and Pattern Recognition}, pp.~1110--1118, 2015.

\bibitem{wang2013action}
H.~Wang and C.~Schmid, ``Action recognition with improved trajectories,'' in
  {\em Computer Vision (ICCV), 2013 IEEE International Conference on},
  pp.~3551--3558, IEEE, 2013.

\bibitem{jiang2015human}
Y.-G. Jiang, Q.~Dai, W.~Liu, X.~Xue, and C.-W. Ngo, ``Human action recognition
  in unconstrained videos by explicit motion modeling,'' {\em IEEE Transactions
  on Image Processing}, vol.~24, no.~11, pp.~3781--3795, 2015.

\bibitem{jain2013better}
M.~Jain, H.~Jegou, and P.~Bouthemy, ``Better exploiting motion for better
  action recognition,'' in {\em Computer Vision and Pattern Recognition (CVPR),
  2013 IEEE Conference on}, pp.~2555--2562, IEEE, 2013.

\bibitem{xu2015learning}
D.~Xu, E.~Ricci, Y.~Yan, J.~Song, and N.~Sebe, ``Learning deep representations
  of appearance and motion for anomalous event detection,'' {\em arXiv preprint
  arXiv:1510.01553}, 2015.

\bibitem{yu2016representation}
J.~Yu, S.~Park, S.~Lee, and M.~Jeon, ``Representation learning, scene
  understanding, and feature fusion for drowsiness detection,'' in {\em
  Computer Vision -- ACCV 2016 Workshops} (C.-S. Chen, J.~Lu, and K.-K. Ma,
  eds.), (Cham), pp.~165--177, Springer International Publishing, 2017.

\bibitem{hong2016learning}
S.~Hong, J.~Oh, H.~Lee, and B.~Han, ``Learning transferrable knowledge for
  semantic segmentation with deep convolutional neural network,'' in {\em
  Proceedings of the IEEE Conference on Computer Vision and Pattern
  Recognition}, pp.~3204--3212, 2016.

\bibitem{zhang2016learning}
Z.~Zhang, P.~Luo, C.~C. Loy, and X.~Tang, ``Learning deep representation for
  face alignment with auxiliary attributes,'' {\em IEEE transactions on pattern
  analysis and machine intelligence}, vol.~38, no.~5, pp.~918--930, 2016.

\bibitem{lecun1998gradient}
Y.~LeCun, L.~Bottou, Y.~Bengio, and P.~Haffner, ``Gradient-based learning
  applied to document recognition,'' {\em Proceedings of the IEEE}, vol.~86,
  no.~11, pp.~2278--2324, 1998.

\bibitem{krizhevsky2012imagenet}
A.~Krizhevsky, I.~Sutskever, and G.~E. Hinton, ``Imagenet classification with
  deep convolutional neural networks,'' in {\em Advances in neural information
  processing systems}, pp.~1097--1105, 2012.

\bibitem{ji20133d}
S.~Ji, W.~Xu, M.~Yang, and K.~Yu, ``3d convolutional neural networks for human
  action recognition,'' {\em IEEE transactions on pattern analysis and machine
  intelligence}, vol.~35, no.~1, pp.~221--231, 2013.

\bibitem{tran2015learning}
D.~Tran, L.~Bourdev, R.~Fergus, L.~Torresani, and M.~Paluri, ``Learning
  spatiotemporal features with 3d convolutional networks,'' in {\em 2015 IEEE
  International Conference on Computer Vision (ICCV)}, pp.~4489--4497, IEEE,
  2015.

\bibitem{le1990handwritten}
B.~B. Le~Cun, J.~S. Denker, D.~Henderson, R.~E. Howard, W.~Hubbard, and L.~D.
  Jackel, ``Handwritten digit recognition with a back-propagation network,'' in
  {\em Advances in neural information processing systems}, Citeseer, 1990.

\bibitem{memisevic2013learning}
R.~Memisevic, ``Learning to relate images,'' {\em IEEE transactions on pattern
  analysis and machine intelligence}, vol.~35, no.~8, pp.~1829--1846, 2013.

\bibitem{hong2015learning}
S.~Hong, J.~Oh, B.~Han, and H.~Lee, ``Learning transferrable knowledge for
  semantic segmentation with deep convolutional neural network,'' {\em arXiv
  preprint arXiv:1512.07928}, 2015.

\bibitem{xu2015show}
K.~Xu, J.~Ba, R.~Kiros, K.~Cho, A.~Courville, R.~Salakhutdinov, R.~S. Zemel,
  and Y.~Bengio, ``Show, attend and tell: Neural image caption generation with
  visual attention,'' {\em arXiv preprint arXiv:1502.03044}, vol.~2, no.~3,
  p.~5, 2015.

\bibitem{abtahi2014yawdd}
S.~Abtahi, M.~Omidyeganeh, S.~Shirmohammadi, and B.~Hariri, ``Yawdd: A yawning
  detection dataset,'' in {\em Proceedings of the 5th ACM Multimedia Systems
  Conference}, pp.~24--28, ACM, 2014.

\bibitem{farfade2015multi}
S.~S. Farfade, M.~J. Saberian, and L.-J. Li, ``Multi-view face detection using
  deep convolutional neural networks,'' in {\em Proceedings of the 5th ACM on
  International Conference on Multimedia Retrieval}, pp.~643--650, ACM, 2015.

\bibitem{parkhi2015deep}
O.~M. Parkhi, A.~Vedaldi, and A.~Zisserman, ``Deep face recognition.,'' in {\em
  BMVC}, vol.~1, p.~6, 2015.

\bibitem{donahue2015long}
J.~Donahue, L.~Anne~Hendricks, S.~Guadarrama, M.~Rohrbach, S.~Venugopalan,
  K.~Saenko, and T.~Darrell, ``Long-term recurrent convolutional networks for
  visual recognition and description,'' in {\em Proceedings of the IEEE
  conference on computer vision and pattern recognition}, pp.~2625--2634, 2015.

\bibitem{parkdriver}
S.~Park, F.~Pan, S.~Kang, and C.~D. Yoo, ``Driver drowsiness detection system
  based on feature representation learning using various deep networks,''

\bibitem{he2015convolutional}
K.~He and J.~Sun, ``Convolutional neural networks at constrained time cost,''
  in {\em Computer Vision and Pattern Recognition (CVPR), 2015 IEEE Conference
  on}, pp.~5353--5360, IEEE, 2015.

\bibitem{notchenko2016sparse}
A.~Notchenko, E.~Kapushev, and E.~Burnaev, ``Sparse 3d convolutional neural
  networks for large-scale shape retrieval,'' {\em arXiv preprint
  arXiv:1611.09159}, 2016.

\end{thebibliography}


%







\newpage

\begin{IEEEbiography}[{\includegraphics[width=1in,height=1.25in,clip,keepaspectratio]{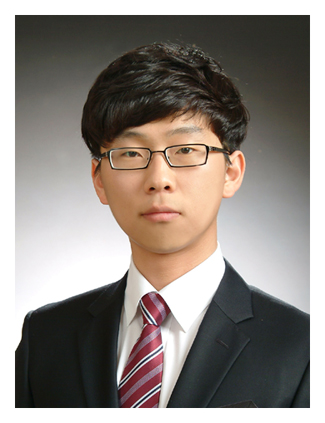}}]{Jongmin Yu} received the B.S. degree in computer science from the Chungnam National University, Daejeon, Republic of Korea, in 2013, and he is a Ph.D candidate of Department of  Electrical Engineering and Computer Science in Gwangju Institute of Science and Technology (GIST), Gwangju, Korea, Republic of. Presently, His research interests include Artificial Intelligence and Machine Learning, Computer Vision, and mathematical understanding and their applications.
	
\end{IEEEbiography}


\begin{IEEEbiography}[{\includegraphics[width=1in,height=1.25in,clip,keepaspectratio]{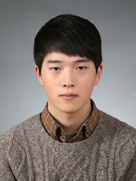}}]{Sangwoo Park} received the B.S. degree in electronic engineering from the Soongsil University, Seoul, Republic of Korea, in 2016, and he is a M.S student of Department of Electrical Engineering and Computer Science in Gwangju Institute of Science and Technology (GIST), Gwangju, Korea, Republic of. Presently, His major interests include Machine Learning, Computer Vision  and mathematical understanding and application of these.
	
\end{IEEEbiography}

\begin{IEEEbiography}[{\includegraphics[width=1in,height=1.25in,clip,keepaspectratio]{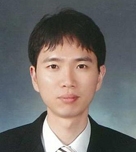}}]{Sangwook Lee}
	
	was born in Busan, Republic of Korea, in 1977. He received the B.E. degree in mechanical engineering from the Korea Advanced Institute of Science and Technology (KAIST), Daejeon, Republic of Korea, in 2000, and the M.S. and Ph.D. degrees in mechatronics engineering from the Gwangju Institute of Science and Technology (GIST), Gwangju, Republic of Korea, in 2002 and 2007, respectively.
	
	In 2007, he joined the School of Computational Science and Engineering, Georgia Institute of Technology, Atlanta, GA, as a postdoctoral researcher. In 2008, he worked at the Communication Research, Samsung electronics, Suwon, Republic of Korea. Since March 2009, he has been with the School of Information and Communication Convergence Engineering, Mokwon University, Daejeon, Republic of Korea, where he was an Assistant Professor, became an Associate Professor in 2015. His current research interests include meta heuristics, combinatorial optimization, artificial intelligence, machine learning, and data clustering.
	
\end{IEEEbiography}

\begin{IEEEbiography}[{\includegraphics[width=1in,height=1.25in,clip,keepaspectratio]{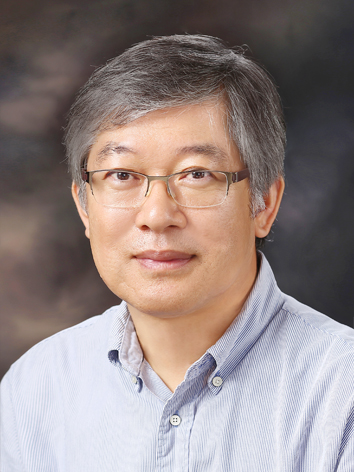}}]{Moongu Jeon}
	
	received the B.S. degree in architectural engineering from the Korea University, Seoul, Korea, in 1988 and the M.S. and Ph.D. degrees
	
	in computer science and scientific computation from the University of Minnesota, Minneapolis, MN, USA, in 1999 and 2001, respectively. As a postgraduate researcher, he worked on optimal control problems at the University of California at Santa Barbara, Santa Barbara, CA, USA, in 2001 to 2003, and then moved to the National Research Council of Canada, where he worked on the sparse representation of high-dimensional data and the image processing until July 2005. In 2005, he joined the Gwangju Institute of Science and Technology, Gwangju, Korea, where he is currently an full professor in the School of Electrical Engineering and Computer Science. His current research interests are in machine learning,computer vision and artificial intelligence.
	
\end{IEEEbiography}


\end{document}